\documentclass[journal]{IEEEtran}

\usepackage{cite}
\usepackage[T1]{fontenc}

\usepackage{float}
\usepackage{graphicx}
\usepackage{subfig}
\usepackage{overpic}

\usepackage{amsthm,amsmath,amssymb}
\usepackage{bm}

\usepackage[ruled,vlined,linesnumbered]{algorithm2e}

\usepackage{array}
\usepackage{booktabs}
\usepackage{enumerate}

\usepackage{url}
\usepackage{color}
\usepackage{multirow}

\begin{document}
%
\title{Parallel Self-assembly for Modular USVs with Diverse Docking Mechanism Layouts}
%
%
%
\author{Lianxin~Zhang, Yang Jiao, Yihan~Huang, Ziyou~Wang, and~Huihuan~Qian 
\thanks{This paper is partially supported by Project U1613226 from NSFC, Project AC01202101105 from the Shenzhen Institute of Artificial Intelligence and Robotics for Society (AIRS), and Project PF.01.000143 from The Chinese University of Hong Kong, Shenzhen, China.}
\thanks{Lianxin Zhang and Huihuan Qian are with Shenzhen Institute of Artificial Intelligence and Robotics for Society (AIRS), The Chinese University of Hong Kong, Shenzhen (CUHK-Shenzhen), Shenzhen, Guangdong, 518129, China. }
\thanks{Lianxin Zhang, Yang Jiao, Yihan Huang, and Huihuan Qian are also with School of Science and Engineering (SSE), The Chinese University of Hong Kong, Shenzhen (CUHK-Shenzhen), Shenzhen, Guangdong 518172, China.}
\thanks{\emph{*Corresponding author is Huihuan Qian (e-mail: hhqian@cuhk.edu.cn).}}
}

\markboth{Transactions on Automation Science and Engineering} 
{Shell \MakeLowercase{\textit{et al.}}: Bare Demo of IEEEtran.cls for IEEE Journals}

\maketitle

\begin{abstract}

Self-assembly enables multi-robot systems to merge diverse capabilities and accomplish tasks beyond the reach of individual robots. 
Incorporating varied docking mechanisms layouts (DMLs) can enhance robot versatility or reduce costs.
However, assembling multiple heterogeneous robots with diverse DMLs is still a research gap.
This paper addresses this problem by introducing CuBoat, an omnidirectional unmanned surface vehicle (USV). CuBoat can be equipped with or without docking systems on its four sides to emulate heterogeneous robots. We implement a multi-robot system based on multiple CuBoats.
To enhance maneuverability, a linear active disturbance rejection control (LADRC) scheme is proposed. 
Additionally, we present a generalized parallel self-assembly planning algorithm for efficient assembly among CuBoats with different DMLs.
Validation is conducted through simulation within 2 scenarios across 4 distinct maps, demonstrating the performance of the self-assembly planning algorithm.
Moreover, trajectory tracking tests confirm the effectiveness of the LADRC controller. Self-assembly experiments on 5 maps with different target structures affirm the algorithm's feasibility and generality. 
This study advances robotic self-assembly, enabling multi-robot systems to collaboratively tackle complex tasks beyond the capabilities of individual robots.

\emph{Note to Practitioners-}
This paper explores the utilization of self-assembly technologies for modular robots with diverse DMLs to facilitate collective construction tasks.
In practical scenarios, swarm robots may engage in various tasks, leading to the adoption of different docking systems or DMLs.
Addressing this variability not only facilitates the integration of a more diverse array of robot systems into self-assembly processes but also has the potential to unveil methods for constructing large structures with fewer docking systems, thereby reducing costs
Achieving efficient task coordination and robot navigation is crucial for assembling substantial quantities of heterogeneous robots.
The proposed method ensures that participating robots can dynamically navigate online, resulting in the successful assembly of a securely connected structure.
This research is particularly relevant for those interested in the effective assembly of heterogeneous robot systems.
Our simulation and hardware experiments validate a conceptual system, affirming the practicability of the algorithm within the multi-robot system.
It's important to note that the introduced approach is not suitable for robots with heterogeneous shapes, three-dimensional target structures, or environments with obstacles.

\end{abstract}


%
\IEEEpeerreviewmaketitle

\section{Introduction}
\label{sect:intro}

 

Modular self-assembly empowers swarm robots for missions beyond single robot capabilities, e.g., space station reconfiguration \cite{RN1545}, intelligent transportation \cite{park2021social}, and collaborative construction \cite{petersen2019review}. 
In these multi-robot systems, individual robots exhibit distinct configurations, encompassing shapes \cite{salvi2017assembly,liu2019magnetically}, functions \cite{lonvcar2019heterogeneous,narvaez2020autonomous}, and mobility \cite{Liu2014Self,yi2022configuration}.
Through self-assembly, heterogeneous modular robot systems can integrate the capabilities of various robots.
In this paper, we are interested in the self-assembly of multi-robot systems with varied docking mechanism layouts (DMLs), which provide at least two advantages over those with identical DMLs.
One is the versatility of individual robots in diverse tasks, facilitated by multiple docking system installation options, allowing for specialized system designs.
Second, reducing the number of required docking systems for self-assembly will decrease manufacturing costs.
However, this presents new challenges for self-assembly planning (SAP).

\begin{figure}
	\centering
	\includegraphics[width=1\linewidth]{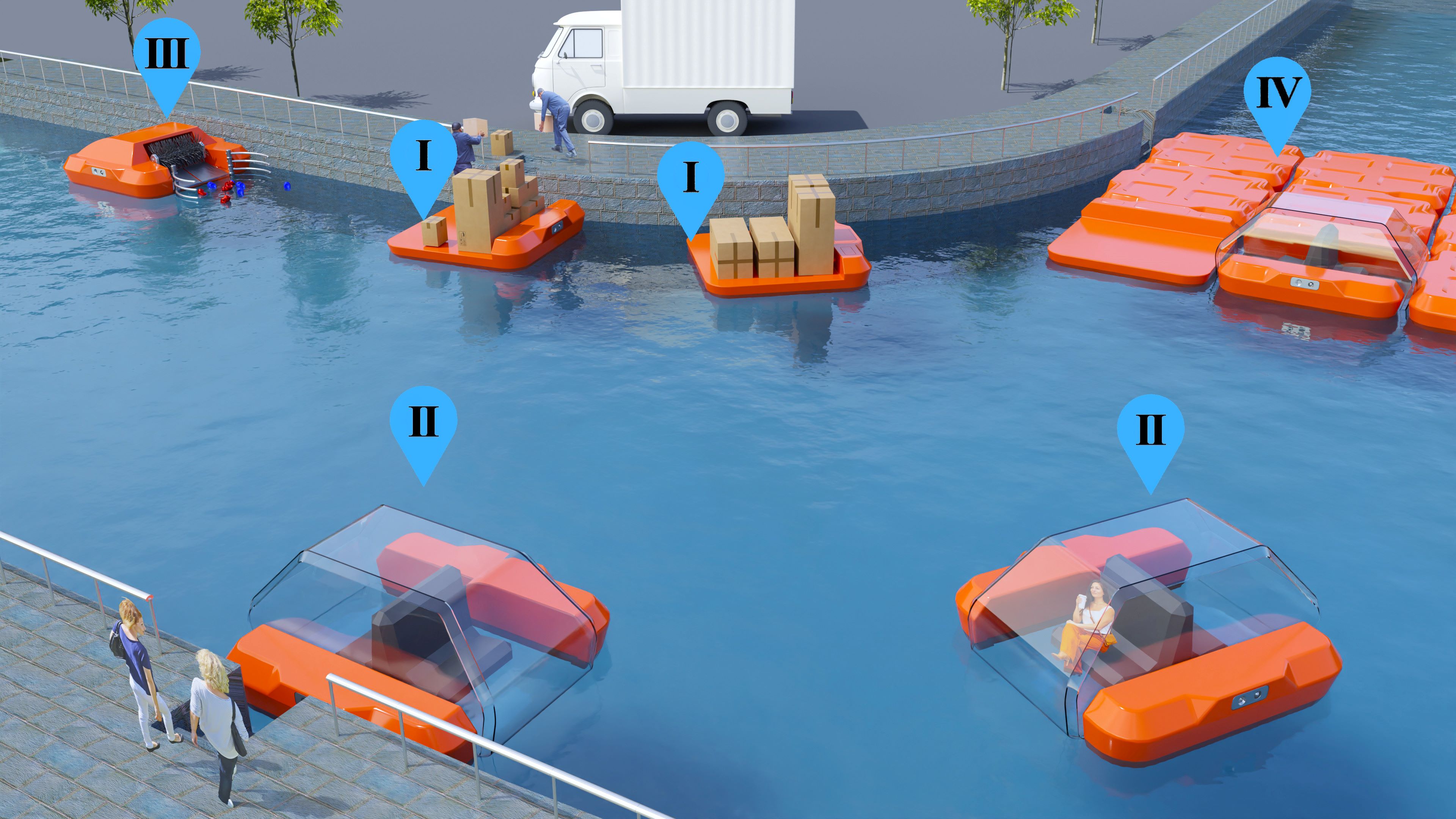} 
	\caption{
		A heterogeneous multi-USV system consists of four types of USVs designed for diverse applications, including transportation and river cleaning.
		These USVs, featuring distinct docking mechanism layouts, can seamlessly assemble into on-demand dynamic platforms.
 }
	\label{fig:design_rendering}
\end{figure}



To investigate the challenges posed by diverse DMLs, we envisage a multi-unmanned surface vehicle (USV) system, acting as dynamic platforms on water surface. 
Our research is motivated by the potential application of self-assembly technologies on water surfaces \cite{wang2015large}.
As shown in Fig. \ref{fig:design_rendering}, different types of omnidirectional USVs with diverse DMLs are incorporated, ranging from Type I to IV.
These USVs can independently undertake various tasks and also connect to form on-demand dynamic floating platforms, like temporary bridges or overwater parking lots. 
To study the effect of heterogeneity in the USVs' docking capability, we develop a scaled heterogeneous USV, named CuBoats, with replaceable docking mechanisms and propose a generalized SAP algorithm to enable rapid self-assembly of CuBoats, based on the PAA method \cite{saldana2017decentralized}. 
Our SAP algorithm can lead the robots with gendered or genderless docking mechanisms to assemble in a parallel manner, which contains four stages: (i) dispatching of target locations, (ii) generating the assembly tree, (iii) extending the target structure, and (iv) navigating the robots to construct the desired formation.
Notably, the SAP algorithm has evolved from self-assembly of homogeneous systems and has been adapted to cater to both homogeneous and heterogeneous systems.

The major contributions we made in this paper are as follows.
\begin{itemize}
	\item A multi-robot system of CuBoats capable of assembling floating structures, where each CuBoat can be equipped with optional docking systems on its four sides. An identical navigation approach is deployed to ensure successful assembly in distributed systems.
	
	\item A novel parallel SAP algorithm is proposed to guide the assembly of the heterogeneous system. The dispatching stage involves solving a non-convex optimization problem constrained by the connectivity of the target structure, which is addressed using the tabu search method.
	
	\item Validation of the proposed SAP algorithm is conducted through simulations involving robots with diverse DMLs on distinct maps.
	
	\item The self-assembly capability of the hardware and the efficiency of the algorithm are demonstrated through 10 sets of experiments using our multi-USV system.
\end{itemize}


The paper is structured as follows. 
Section \ref{sect:relatedworks} reviews related works. Section \ref{sect:design} presents the components and control system of the multi-USV system. In Section \ref{sect:problem}, the concerned SAP problem is introduced. The proposed SAP algorithm and its detailed steps are outlined in Section \ref{sect:algorithm}. Section \ref{sect:evaluation} describes simulations conducted on four robots with different DMLs. The validation experiment results, including tracking and assembly, are discussed in Section \ref{sect:experiment}. Finally, Section \ref{sect:conclusion} provides the paper's conclusion.

\section{Related Works}
\label{sect:relatedworks}

Docking between heterogeneous robots is a common requirement in cross-domain multi-robot systems, such as air-ground robot teams \cite{lonvcar2019heterogeneous} and surface-underwater robot swarms \cite{narvaez2020autonomous}.
Cooperation among these diverse robots allows them to sense the environment from different perspectives and perform tasks with improved performance.
While self-assembly technology has been extensively studied for robots with diverse shapes \cite{salvi2017assembly,liu2019magnetically}, different mobility \cite{Liu2014Self,yi2022configuration}, and robot-block connections in collective construction \cite{werfel2007collective, werfel2008three}, the specific challenge posed by heterogeneous docking mechanisms has been addressed in some works \cite{gross2006autonomous}.
Nevertheless, there remains a lack of studies involving multi-USV systems with different DMLs.


In past decades, numerous SAP algorithms have been developed to ensure collision-free paths and correct assembly sequences, falling into two categories: serial and parallel approaches.
Initially, the serial docking approach designates one robot as the seed to which others connect sequentially, e.g., the seed-initiated rule-based methods \cite{baldassarre2007self, haire2019ship}, the chain forming approach \cite{RN1066}, and the collective robotic construction \cite{werfel2007collective, werfel2008three}, but its time consumption increases linearly with the number of robots.
Researchers have attempted to accelerate this approach by parallelizing robot aggregation into one connected component, using methods like growing multiple branches from one seed \cite{knychalatucci:hal-02182793, CL:QL:HK:MY:20}, setting up multiple seeds \cite{seo2013assembly, jilek2022self}, or moving two chains individually \cite{yang2019distributed}.
However, the number of branches does not scale with the robots, making it inefficient for large-scale applications.
While some algorithms, inspired by passive self-assembly, introduced concurrent assembly processes \cite{klavins2006grammatical, klavins2007programmable} and centroidal Voronoi tessellation-based approaches \cite{wei2017centroidal} where robots move synchronously by ignoring collisions and obstacles.
To avoid collisions and undesired attachments, a fully-parallel self-assembly algorithm with a binary assembly tree, named PAA, was proposed \cite{saldana2017decentralized}, and further work involved avoidance of immovable obstacles \cite{zhang2021efficient,Zhang2023Parallel}.
However, existing parallel algorithms are solely designed for homogeneous systems.

In this paper, we propose a heterogeneous multi-USV system consisting of CuBoats to mimic practical robots with diverse DMLs. 
The CuBoats can be equipped with either genderless or gender-opposite docking systems, leading to two distinct scenarios addressed by the SAP algorithm.
We also address the limitation of existing SAP algorithms by introducing an extended concept of parallel docking that accommodates either homogeneous or heterogeneous systems with diverse docking systems.
Efficient dispatching planning is provided to ensure the connectivity of the structure formed by modular robots.
The proposed SAP algorithm is subjected to simulations, and experiments are performed on the multi-USV system to validate its robustness and efficiency.

\begin{figure} [htbp] 
	\centering
	\includegraphics[width=1\linewidth]{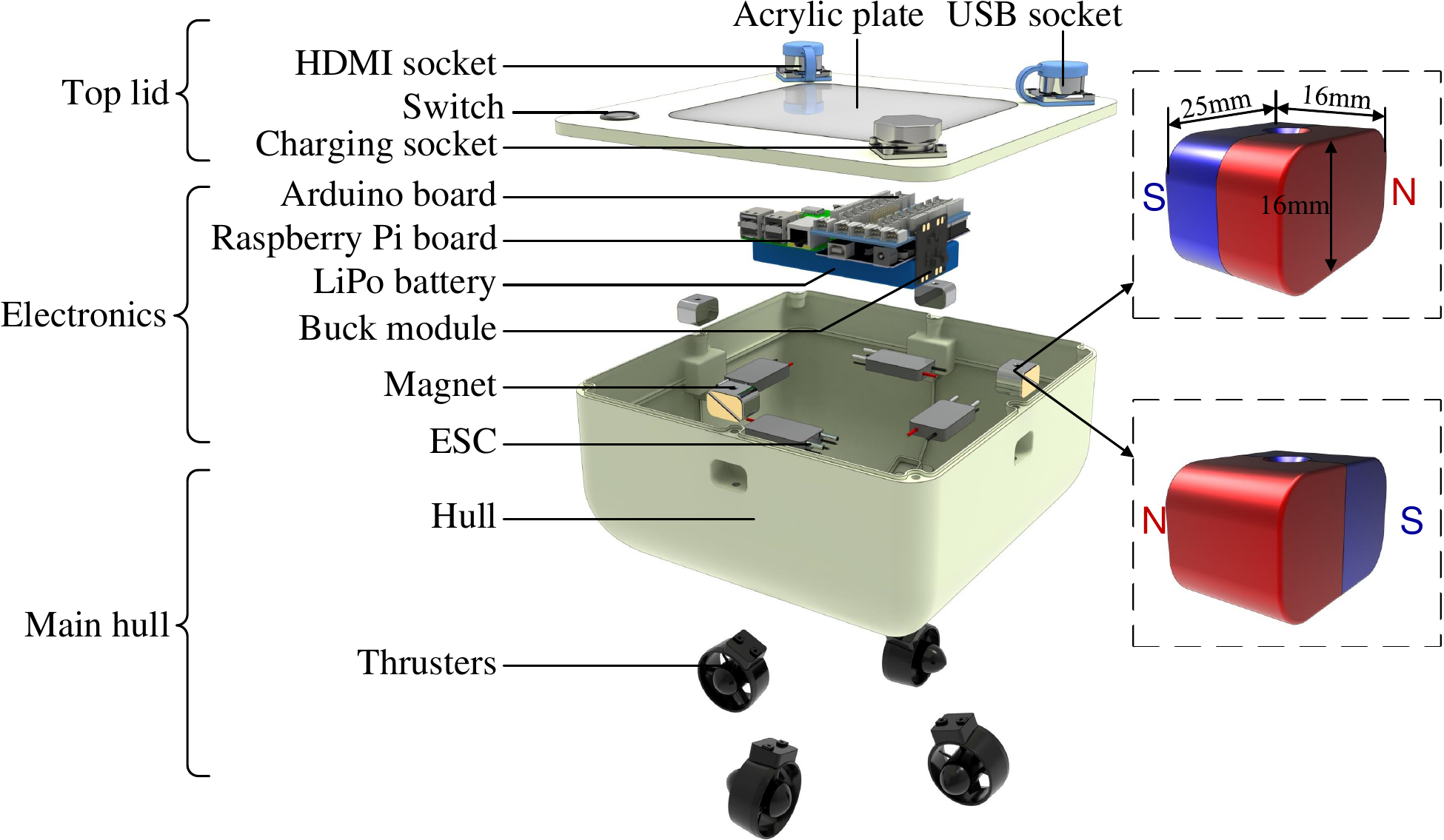}
	\caption{System overview of the CuBoat. }
	\label{fig:cuboat}
\end{figure}

\section{MODULAR ROBOTS AND CONTROL METHOD}
\label{sect:design}


\subsection{Design and Fabrication of CuBoat} 
\label{sect:USV}
Fig. \ref{fig:cuboat} presents the main components of the individual CuBoat design, which includes the hull, docking magnets, propulsion system, and electronics. Each component is detailed in the following sections.

\subsubsection{Hull and propulsion system} 

The CuBoat features a cube-shaped design with holonomic propulsion enabled by four thrusters placed circumferentially on the bottom. Its interior houses the electronics, including the Raspberry Pi board, and magnets. The lid is reserved for the switch, as well as sockets dedicated to charging and data transmission.

\begin{figure} [htbp] 
	\centering
	\includegraphics[width=0.7\linewidth]{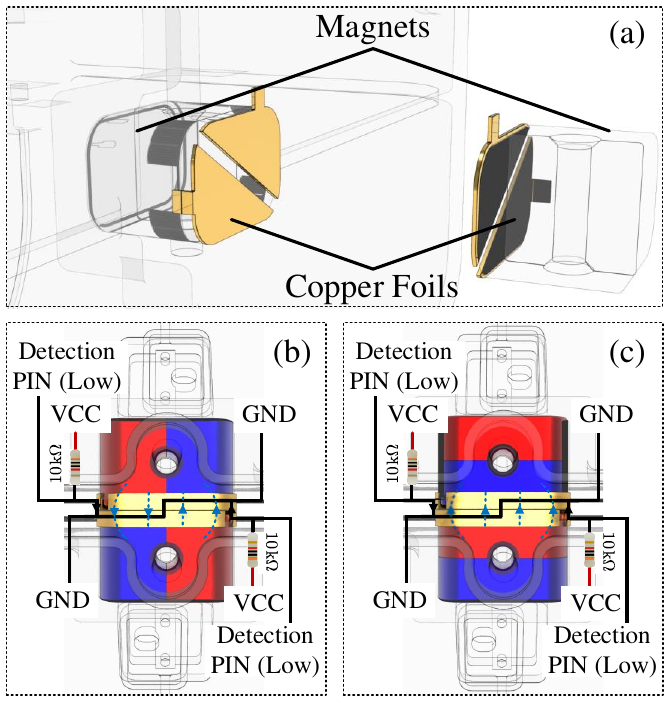}
	\caption{(a) The undocked state of the docking subsystem. The docked state of (b) the genderless pattern and (c) the gendered pattern. The contact detection mechanism is presented.}
	\label{fig:DockingSystem}
\end{figure}

The CuBoat's cubic hull and top lid are separately 3D printed using acrylonitrile butadiene styrene. 
Each hull wall on the four sides can accommodate a replaceable permanent magnet, as shown in Fig. \ref{fig:DockingSystem} (a). 
These magnets generate a powerful attraction force of up to 112 N when two CuBoats are brought close, ensuring a strong and secure connection between them.

\begin{figure} [htbp] 
	\centering
	\includegraphics[width=0.7\linewidth]{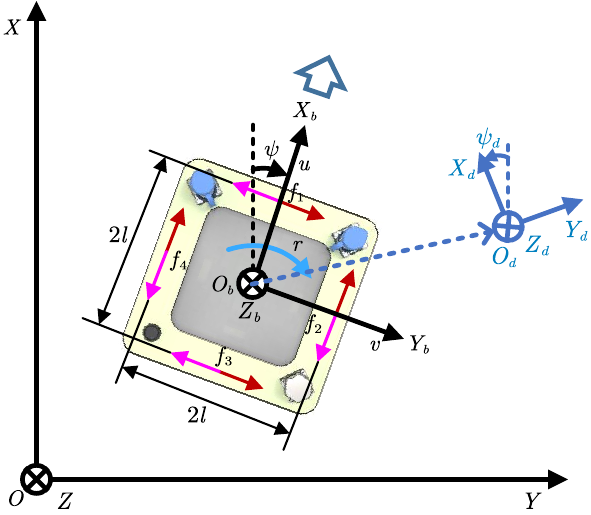}
	\caption{Coordinate system for the movement of the CuBoat: the inertial coordinate $O \text{-} X Y Z$, the body coordinate $O_b \text{-} X_b Y_b Z_b$ and the target coordinate $O_d \text{-} X_d Y_d Z_d$. Red arrows represent positive propulsion forces, while pink arrows indicate negative forces.}
	\label{fig:motion_control}
\end{figure}

The CuBoat is equipped with four thrusters arranged in a "+" shaped configuration, which has been proven in \cite{Zhang2023Parallel} more efficient for propulsion than other configurations such as the "X" shaped configuration \cite{NAD2015172} or the three-thruster configuration \cite{vallegra2018gradual}. In Fig. \ref{fig:motion_control}, the body coordinate system and thruster layout are depicted, with all thrusters capable of generating both forward and backward forces. The propulsion forces generated by the four thrusters are denoted as $f_i \ (i=1,2,3,4)$, and $L_i$ represents the distance from each thruster to the body center. The applied force and moment vector $\bm{\tau}$ in the plane can be represented as
\begin{equation} \label{eq:forces}
\bm{\tau } = \mathbf{E}\bm{u} =
\left[ \begin{matrix}
0&		1&		0&		1\\
1&		0&		1&		0\\
l&	   -l&	   -l&	    l\\
\end{matrix} \right] \left[ \begin{array}{c}
f_1\\
f_2\\
f_3\\
f_4\\ 
\end{array} \right].
\end{equation}

\subsubsection{Electronics} 
%
The CuBoat's main electronic components consist of sensors, a processor with control circuits, and the power supply.
For localization, the OptiTrack motion capture system is employed, providing millimeter-precision positions through a local area network (LAN) and a virtual-reality peripheral network (VRPN) interface, which connects to the CuBoats.
%
To monitor the docking state during the assembly process, a contact detection circuit is implemented. The schematic diagram of this circuit is depicted in Fig. \ref{fig:DockingSystem} (b) and (c).
%
The circuit employs two pieces of copper foil, segmented by oblique cutting, to connect to the detection pin and the ground, respectively. This arrangement ensures that the foils on the two sides are mutually closed during the connection.
After docking, the magnetic force presses the two-side foils together, causing both detection pins to connect with the ground. The voltage of the detection pin is high when two USVs are separated and low when two USVs are docked, effectively indicating the docking status.

%
The processing unit of the CuBoat utilizes a Raspberry Pi 4B board (1.5 GHz ARM Cortex-A72, 2G LPDDR4) running the robot operating system (ROS) to execute navigation and control algorithms. 
To efficiently manage tasks, an Arduino board is employed to directly control the four thrusters of the propulsion subsystem, enabling the processor to focus on communication and higher-level control.
For power supply, a LiPo battery with a capacity of $9800 \rm mAh$, coupled with a buck module, provides both 5V and 12V power. This setup endows the CuBoat with a battery life of approximately 2 hours.

\subsection{Modeling} 
\label{sect:model}
To establish the equations of motion, we represent the CuBoat in a three-dimensional inertial coordinate system ($O_w \text{-} X_wY_wZ_w$), as shown in Fig. \ref{fig:motion_control}.
For planar motion on the water surface, we define the location and orientation of the robot as $\boldsymbol{\eta}=\left[ x\, y\, \psi \right]^T$.
Additionally, a body-fixed coordinate system ($O_b \text{-} X_bY_bZ_b$) is set at the body center of the CuBoat, with its origin approximately at the center of mass (COM) due to structural symmetry.
The $O_b \text{-} X_b$ axis coincides with the longitudinal axis (from aft to fore).
In the body frame, the robot velocity is denoted as $ \bm{v} =\left[ u\,\,v\,\,r \right] ^T$.

Without the presence of disturbances, the kinematic model can be expressed by
\begin{equation} \label{eq:kinematic}
\boldsymbol{\dot{\eta}}=\mathbf{R}\left( \psi \right) \bm{v} ,
\end{equation}
where $\mathbf{R}\left( \psi \right)$ is the transformation matrix converting a state vector from body frame to inertial frame, denoted as
\begin{equation} \label{eq:transform}
\mathbf{R}\left( \psi \right) =\left[ \begin{matrix}
\cos \psi&		-\sin \psi&		0\\
\sin \psi&		\cos \psi&		0\\
0&		0&		1\\
\end{matrix} \right]. 
\end{equation}

According to the Fossen model in \cite{fossen2011handbook}, the dynamic model of the TransBoat can be represented as
\begin{equation} \label{eq:dynamic}
\mathbf{M} \dot{\bm{v}}+\mathbf{C}\left( \bm{v} \right) \bm{v}+ \mathbf{D }\left( \bm{v} \right) \bm{v}=\bm{\tau} + \bm{\tau}_w,
\end{equation}
where $\bm{\tau}$ is the force and moment vector defined in Eq. (\ref{eq:forces}) and  $\bm{\tau}_w$ denotes the forces and moments induced by the disturbances, such as wind and waves. The coefficients are defined as follows. First, $\mathbf{ M}$ is a diagonal mass matrix combining the robot's inertial mass and the added mass, since
the origin $O_b$ coincides with the COM, that is
\begin{equation}
\mathbf{M} = \textrm{diag} \left\{ m_{1}, m_{2}, m_{3} \right\} .
\end{equation}
Second, $\mathbf{C}\left( \bm{v} \right)$ is the matrix of Coriolis and centripetal terms
\begin{equation}
\mathbf{C}\left(\bm{v} \right) =
\left[ \begin{matrix}
0&		0&		-m_{2} v\\
0&		0&		m_{1} u\\
m_{2} v&		-m_{1} u&		0\\
\end{matrix} \right]. 
\end{equation}
Third, considering the low movement speed and the symmetrical structure, we adopt the linear drag matrix $\mathbf{D}$ as
\begin{equation}
\mathbf{D} =
\textrm{diag} \{X_u,  Y_v, N_r \}.
\end{equation}

\begin{figure*} [htpb]
	\centering
	\includegraphics[width=1\linewidth]{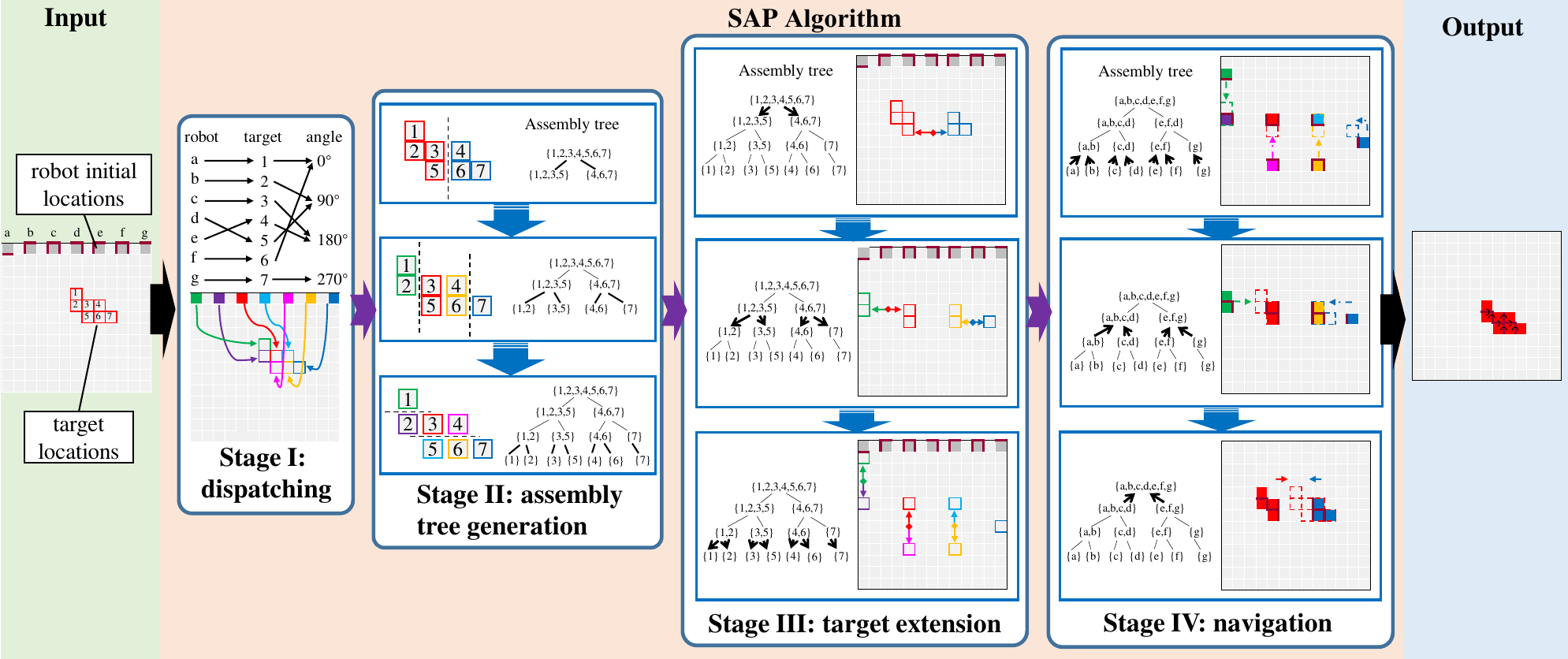}
	\caption{Overview of the SAP algorithm. A seven-robot self-assembly process illustrates the four stages vividly. Homogeneous docking systems are indicated by the bold red borders of robot points. For stages II and IV, subplots in the panels depict the recording and tracing processes of the landmark points, respectively.	}
	\label{fig:pipeline}
\end{figure*}

\subsection{Control} 
\label{sect:control}

From Eq. (\ref{eq:kinematic}), the velocity in body frame is 
\begin{equation}
\bm{v}=\mathbf{R}^{-1}\left( \psi \right) \boldsymbol{\dot{\eta}}.
\end{equation}
Substitute it into Eq. (\ref{eq:dynamic}), and we obtain the second-order dynamic equation as
\begin{equation} \label{eq:adrc_dynamics}
\boldsymbol{\ddot{\eta}}=\mathbf{f}\left( \boldsymbol{\eta },\boldsymbol{\dot{\eta}}, \boldsymbol{\tau }_w \right) +\mathbf{B}\bm{u},
\end{equation}
where $\mathbf{B}=\mathbf{RM}^{-1}\mathbf{E}$ and the nonlinear term including the disturbance is 
\begin{equation} \label{eq:nonlinear_term}
\mathbf{f}\left( \boldsymbol{\eta },\boldsymbol{\dot{\eta}}, \boldsymbol{\tau }_w \right) = \mathbf{RM}^{-1}\boldsymbol{\tau }_w-\mathbf{RM}^{-1}\left( \mathbf{C} +\mathbf{D} \right) \mathbf{R}^{-1}\boldsymbol{\dot{\eta}}-\mathbf{R\dot{R}}^{-1}\mathbf{\boldsymbol{\eta}}.
\end{equation}
This nonlinear term constantly changes with the system state and control input, requiring identification.

Let the state variables become
$\bm{x}_1=\boldsymbol{\eta}, \bm{x}_2=\boldsymbol{\dot{\eta}}$,
and we introduce two extra variables
\begin{equation}
\bm{x}_3=\mathbf{f}\left( \boldsymbol{\eta },\boldsymbol{\dot{\eta}}, \boldsymbol{\tau }_w \right), \quad
\bm{x}_4=\mathbf{\dot{f}}\left( \boldsymbol{\eta },\boldsymbol{\dot{\eta}}, \boldsymbol{\tau }_w \right).
\end{equation}
The system state is then extended to be
\begin{equation}
\begin{aligned}
\bm{\dot{x}}_1 = \bm{x}_2, \quad
\bm{\dot{x}}_2 = \bm{x}_3+\mathbf{B}\bm{u}, \quad
\bm{\dot{x}}_3 = \bm{h}, \quad
\bm{y} = \bm{x}_1,
\end{aligned}
\end{equation}
which can be rewritten in the compact form
\begin{equation} \label{eq:system_state}
\begin{aligned}
\bm{\dot{x}}&=\widetilde{\mathbf{A}}\bm{x}+\widetilde{\mathbf{B}}\bm{u}+\widetilde{\mathbf{G}}\bm{h}, \\
\bm{y} &= \widetilde{\mathbf{C}}\bm{x},
\end{aligned}
\end{equation}
where $\bm{x}=\left[ \begin{matrix}
\bm{x}_1&		\bm{x}_2&		\bm{x}_3\\
\end{matrix} \right] ^T$ is the extended state vector, and
\begin{equation}
\begin{aligned}
\widetilde{\mathbf{A}} &= \left[ \begin{matrix}
\mathbf{0}_{3\times 3}&		\mathbf{I}_{3\times 3}&		\mathbf{0}_{3\times 3}\\
\mathbf{0}_{3\times 3}&		\mathbf{0}_{3\times 3}&		\mathbf{I}_{3\times 3}\\
\mathbf{0}_{3\times 3}&		\mathbf{0}_{3\times 3}&		\mathbf{0}_{3\times 3}\\
\end{matrix} \right] , 
\,\,\widetilde{\mathbf{B}} = \left[ \begin{matrix}
\mathbf{0}_{3\times 4}&		\mathbf{B}_{3\times 4}&		\mathbf{0}_{3\times 4}\\
\end{matrix} \right] ^T, \\
\mathbf{\tilde{G}} &= \left[ \begin{matrix}
\mathbf{0}_{3\times 3}&		\mathbf{0}_{3\times 3}&		\mathbf{I}_{3\times 3}\\
\end{matrix} \right] ^T, 
\mathbf{\tilde{C}} = \left[ \begin{matrix}
\mathbf{I}_{3\times 3}&		\mathbf{0}_{3\times 6}\\
\end{matrix} \right] .
\end{aligned}
\end{equation}

To observe the coupled three-dimensional motion of the CuBoat in the above system, a linear extended state observer (LESO) is designed as follows.
\begin{equation} \label{eq:LESO}
\begin{aligned}
\bm{\dot{\hat{x}}}
&= \left( \widetilde{\mathbf{A}}-\mathbf{L}\widetilde{\mathbf{C}} \right) \bm{\hat{x}}+\left[ \begin{matrix}
\widetilde{\mathbf{B}}&		\mathbf{L}\\
\end{matrix} \right] \left[ \begin{array}{c}
\bm{u}\\
\bm{y}\\
\end{array} \right] ,
\end{aligned}
\end{equation}
where $\bm{\hat{x}}$ and $\mathbf{L}$ are the state vector and parameter matrix of the observer, respectively. 
Subtracting Eq. (\ref{eq:LESO}) from (\ref{eq:system_state}) and letting $\bm{e}=\bm{x}-\hat{\bm{x}}$, we can express the error dynamics as
\begin{equation} \label{eq:LESO_error}
\dot{\bm{e}}=\left( \widetilde{\mathbf{A}}-\mathbf{L}\widetilde{\mathbf{C}} \right) \bm{e}+\widetilde{\mathbf{G}}\bm{h}.
\end{equation}
We next denote the coefficient matrix $\widetilde{\mathbf{A}}-\mathbf{L}\widetilde{\mathbf{C}}$ by $\mathbf{P}$. The error $\mathbf{e}$ will converge to zero with $\mathbf{P}$ being negative definite. 

The parameter matrix $\mathbf{L}$ contains $27$ elements, making it large-scale for manually tuning. To facilitate the tuning process, we decouple the LESO between different motions by neglecting the coupling elements of $\mathbf{L}$, and therefore let \begin{equation}  \label{eq:parameter_L}
\mathbf{L}=\left[ \begin{matrix}
\mathbf{L}_1&		\mathbf{L}_2&		\mathbf{L}_3\\
\end{matrix} \right] ^T ,
\end{equation}
where $\mathbf{L}_1, \mathbf{L}_2, \mathbf{L}_3$ are $3 \times 3$ diagonal matrices with
\begin{equation}
\begin{aligned}
\mathbf{L}_1 &= \textrm{diag} \left\{ l_{11}, l_{12}, l_{13} \right\} , \\
\mathbf{L}_2 &= \textrm{diag} \left\{ l_{21}, l_{22}, l_{23} \right\} , \\
\mathbf{L}_3 &= \textrm{diag} \left\{ l_{31}, l_{32}, l_{33} \right\}.
\end{aligned}
\end{equation}
Then, we write the simplified matrix $\mathbf{P}$ as
\begin{equation}
\mathbf{P}=\left[ \begin{matrix}
-\mathbf{L}_1&		\mathbf{I}_{3\times 3}&		\mathbf{0}_{3\times 3}\\
-\mathbf{L}_2&		\mathbf{0}_{3\times 3}&		\mathbf{I}_{3\times 3}\\
-\mathbf{L}_3&		\mathbf{0}_{3\times 3}&		\mathbf{0}_{3\times 3}\\
\end{matrix} \right].
\end{equation}
When all parameters $l_1 \sim l_9$ are positive, $\mathbf{P}$ is a Hurwitz matrix which guarantees the convergence of $\bm{e}$.
When the LESO (\ref{eq:LESO}) converges, the system state $\hat{\bm{x}}$ can be observed including the uncertain nonlinear term $\mathbf{f}\left( \boldsymbol{\eta },\boldsymbol{\dot{\eta}}, \boldsymbol{\tau }_w \right)$ in (\ref{eq:nonlinear_term}).


%
Here, we employ the control law as follows,
\begin{equation} \label{eq:adrc_control_law}
\bm{u}=\mathbf{B}^{\dag}\left( \bm{u}_0-\mathbf{\hat{f}}\left( \boldsymbol{\eta },\boldsymbol{\dot{\eta}},\boldsymbol{\tau }_w \right) \right) ,
\end{equation}
where $\mathbf{\hat{f}}\left( \boldsymbol{\eta },\boldsymbol{\dot{\eta}},\boldsymbol{\tau }_w \right)$ is the estimated nonlinear term, $\mathbf{B}^{\dag}=\mathbf{B}^T\left( \mathbf{B}\mathbf{B}^T \right) ^{-1}$ is the pseudo-inverse of $\mathbf{B}$, as $\mathbf{B}$ is full row rank. $\bm{u}_0$ is the nominal control input to be determined.
By substituting Eq. (\ref{eq:adrc_control_law}) into (\ref{eq:adrc_dynamics}), we obtain the dynamics as
\begin{equation}
\boldsymbol{\ddot{\eta}}=\mathbf{f}\left( \boldsymbol{\eta },\boldsymbol{\dot{\eta}},\boldsymbol{\tau }_w \right) +\bm{u}_0-\mathbf{\hat{f}}\left( \boldsymbol{\eta },\boldsymbol{\dot{\eta}},\boldsymbol{\tau }_w \right).
\end{equation}
When $\mathbf{f}$ approaches $\mathbf{\hat{f}}$, the dynamics becomes
\begin{equation}
\boldsymbol{\ddot{\eta}}=\bm{u}_0.
\end{equation}

Here, we can employ a PD controller, that is
\begin{equation}
\bm{u}_0=\mathbf{K}_p\left( \boldsymbol{\eta }_{r}-\boldsymbol{\eta } \right) +\mathbf{K}_d\left( \boldsymbol{\dot{\eta}}_{r}-\boldsymbol{\dot{\eta}} \right)
\end{equation}
with reference states $\boldsymbol{\eta }_{r}$ and $\boldsymbol{\dot{\eta}}_{r}$.
The controller gains $\mathbf{K}_p$ and $\mathbf{K}_d$ are set as
\begin{equation}
\begin{aligned}
\mathbf{K}_p &= \textrm{diag} \left\{ K_{p1}, K_{p2}, K_{p3} \right\} , \\
\mathbf{K}_d &= \textrm{diag} \left\{ K_{d1}, K_{d2}, K_{d3} \right\} ,
\end{aligned}
\end{equation}
which ensure the Hurwitz stability of the ADRC controller \cite{Zheng2009active}.
The LESO parameters $\mathbf{L}$ and the PD gains $\mathbf{K}_p, \mathbf{K}_d$ are exhibited in Table \ref{tab:adrc_gains}.

\begin{table}[htbp]
	\centering
	\caption{ADRC CONTROLLER GAINS}
	\label{tab:adrc_gains}
	\setlength{\tabcolsep}{10pt}{
		\begin{tabular}{cccccc}
			\toprule[1.5pt]
			{$K_{p1}$} & {$K_{p2}$} & {$K_{p3}$} & {$K_{d1}$} & {$K_{d2}$} & {$K_{d3}$} \\
			\hline
			1600      & 1600     & 144       & 80          & 80        & 24 \\
			\midrule[1.5pt]
			{$l_{11}(l_{21})$} & {$l_{12}(l_{22})$} & {$l_{13}(l_{23})$} & {$l_{31}$} & {$l_{32}$} & {$l_{33}$} \\
			\hline
			30      & 300     & 300       & 9          & 27        & 27 \\
			\bottomrule[1.5pt]
	\end{tabular}}
\end{table}

\section{PROBLEM DEFINITION}
\label{sect:problem}

\subsection{Preliminaries and Notations}
\begin{figure} [htpb]
	\centering
	\includegraphics[width=0.8\linewidth]{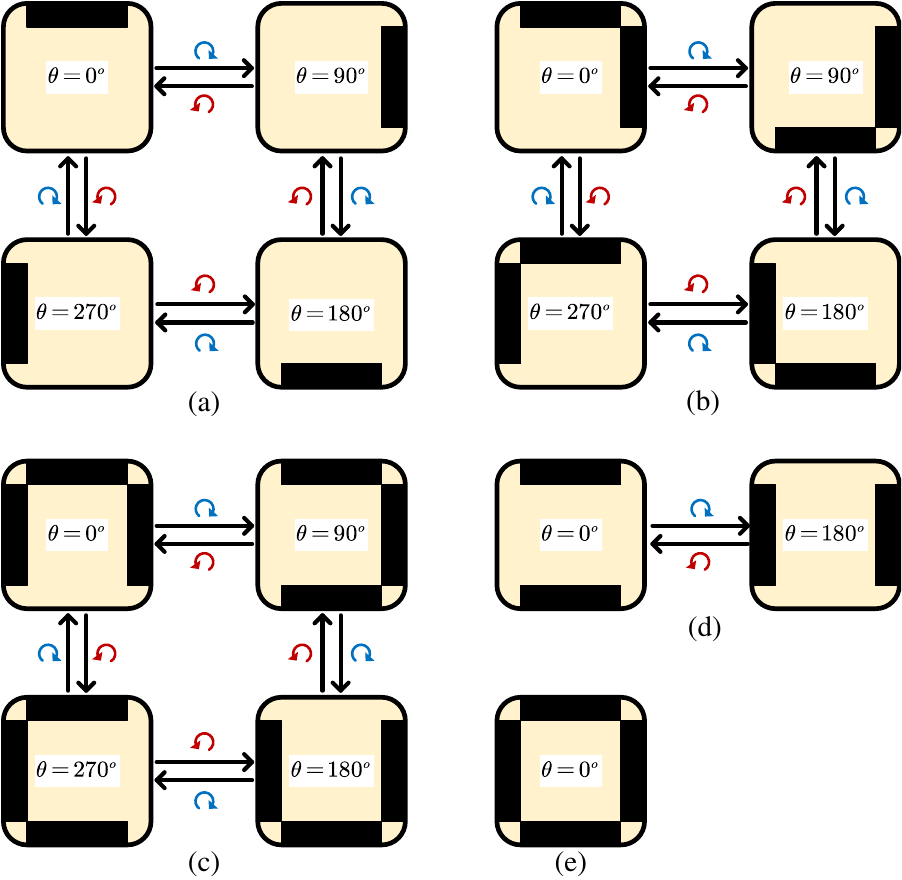}
	\caption{Representation of five DML types and their transitions between different orientation states. A set $\mathbf{\Phi}$ incorporates all these states.}
	\label{fig:dml}
\end{figure}

To facilitate analysis, the CuBoats are abstracted as square modular robots with an identical length $w$.
We have $N$ square omnidirectional robots in the Euclidean space $\mathbb{R}^2$, each equipped with docking systems on their sides (not necessarily all four sides).
These robots are represented by a set  $\mathbf{A} =\left\{ \bm{a}_1,\bm{a}_2,\dots ,\bm{a}_{N} \right\}$, where each $\bm{a}_i$ has a location $\bm{p}_{a_i}\left(t\right) \in \mathbb{R}^2$, an orientation $\psi_{a_i}\left(t\right) \in \mathbb{R}$ at time $t$, and a docking system layout $\delta _{a_i}$.
The robot's DML $\delta _{a_i}$ and orientation $\psi_{a_i}\left(t\right)$ jointly influence its docking action, and their combined state always belongs to one of the states in Fig.  \ref{fig:dml}, represented as a single set $\mathbf{\Phi}$.
The unspecified $M$ target locations are denoted by $\mathbf{G}=\left\{ \bm{g}_1,\bm{g}_2,\dots ,\bm{g}_M \right\} $ with $M \le N$ to ensure that each target can be assigned to a robot.
These dispatched robots are denoted by $\mathbf{A}_g $ with $\mathbf{A}_g\subseteq \mathbf{A}$.
The mapping from robots to targets is expressed through a transformation of the set $\mathbf{A}_g$, denoted as $T\left(\mathbf{A}_g\right)=\left\{ \tilde{\boldsymbol{a}}_1, \tilde{\boldsymbol{a}}_2,\dots ,\tilde{\boldsymbol{a}}_M \right\}$.
Each robot's control inputs, denoted as $\boldsymbol{v}_i \in \mathbb{R}^2$ and $\omega_i \in \mathbb{R}$, adheres to first-order kinematics, i.e., $\dot{\bm{p}}_{a_i} = \bm{v}_i$ and $\dot{\psi}_{a_i} = \omega_i$. 
%
$\lVert \cdot \rVert $ denotes the Euclidean norm.
$\left| \cdot \right|$ denotes the element number, while $\left| \cdot \right|_c$ denotes the connectivity number of a structure.

\subsection{SAP Problem Formulation}

\subsubsection{Given Information}
%
At the beginning, the initial positions $\bm{p}_{a_i}\left(0\right)$, orientations $\psi_{a_i}\left(0\right)$, and docking system layouts $\delta _{a_i}$ for each robot $\bm{a}_i$ in set $A$, along with their corresponding target locations $\mathbf{G}$, are specified.
At the initial time $t_0$, the initial state of the robots are set to $\mathbf{A}\left(t_0\right)=\mathbf{A}_0\left(\bm{p}_{a_i}\left(0\right),\psi_{a_i}\left(0\right),\delta _{a_i}\right)$, as shown in Fig. \ref{fig:pipeline}.

\subsubsection{Constraints}
\begin{itemize}
\item At the end time $t_e$, the resulting structure must be connected to ensure coherence, i.e., $\left| \mathbf{A}_g\left( t_e \right) \right|_c=1$.
\item Robots are required to steer clear of collisions with other robots during their movements, , meaning that $\lVert \bm{p}_{a_i}\left( t \right) -\bm{p}_{a_j}\left( t \right) \rVert _{2}^{2}\geqslant w,\forall i,j\in N,i\ne j$.
\end{itemize}

\subsubsection{Assumptions}
\begin{itemize}
\item All robots move forward/backward/leftward/rightward at a uniform speed of 1 robot length per step and can swerve to  any orientation angle within a time step, namely, $\boldsymbol{v}_i \in \left\{ \left[ \begin{matrix}
0&		0\\
\end{matrix} \right] ^T,\left[ \begin{matrix}
\pm w&		0\\
\end{matrix} \right] ^T,\left[ \begin{matrix}
0&		\pm w\\
\end{matrix} \right] ^T \right\}$ and $\omega_i \in \left(-\pi,\pi\right)$.
\item 
Two robots will connect when their sides with homogeneous docking systems (or opposed sides of gendered docking systems) are adjacent, and this connection will be maintained until the end of the process.
\item 
Docking action can only occur between two groups of robots, subject to environmental disturbances.
\end{itemize}

Our objective is to design a SAP algorithm that guides a fleet of robots with various DMLs to assemble the desired structure at designated target locations.
The primary aim of the algorithm is to minimize the overall number of moving steps needed for navigating the robots while ensuring the connectivity of the formed structure.
Overall, the SAP problem is formulated as
\begin{equation} \label{eq:sap_problem}
\begin{aligned} 
&\underset{T\left( \mathbf{A} \right) ,\boldsymbol{v}_i,\omega_i}{\min}\,\,\int_{t_0}^{t_e}{1dt}
\\
\text{s}.\text{t}. \quad & \mathbf{A}\left( t_0 \right)=\mathbf{A}_0, \quad
\left| \mathbf{A}_g\left( t_e \right) \right|_c=1,
\\
&\sum_{\tilde{\boldsymbol{a}}\in T\left( \mathbf{A} \right)}{\lVert \bm{p}_{\tilde{\boldsymbol{a}}_i}\left( t_e \right)-\boldsymbol{g}_{\tilde{\boldsymbol{a}}} \rVert _{2}^{2}}=0, 
\\
&\dot{\bm{p}}_{a_i} = \bm{v}_i,
\quad
\dot{\psi}_{a_i} = \omega_i,
\\
&\lVert \bm{p}_{a_i}\left( t \right) -\bm{p}_{a_j}\left( t \right) \rVert _{2}^{2}\geqslant w, \forall i,j\in N,i\ne j,
\end{aligned} 
\end{equation}
where $\boldsymbol{g}_{\tilde{\boldsymbol{a}}}\in \mathbf{G}$ is the target location corresponding to the assigned robot $\tilde{\boldsymbol{a}}$.
Without loss of generality, we will examine this problem in a two-dimensional grid map, where each robot resides within one cell, denoted by $w=1$.

\section{Approach}
\label{sect:algorithm}

\subsection{Pipeline}

Building upon the PAA algorithm \cite{saldana2017decentralized} and the assembly-by-extension technique from prior works \cite{zhang2021efficient,Zhang2023Parallel}, we present a parallel SAP algorithm, outlined in Fig. \ref{fig:pipeline}. 
Upon receiving the initial positions, desired locations, and docking configurations of the robots, this algorithm efficiently guides the robots through a sequence of steps. 
Initially, the robots align themselves in the correct orientation, then they proceed to expand towards the target locations, finally assembling the desired structure incrementally.
A key contribution of this paper lies in Stage I of the algorithm, while Stages II and III are derived from state-of-the-art methods. Notably, our algorithm ensures the connectivity of the final structure, ensuring a successful and coherent assembly process.

In Stage I, the allocation of target locations to each robot is formulated as an optimization problem and efficiently solved using the tabu search algorithm \cite{Glover1998Tabu}.
In Stage II, the algorithm systematically divides the target structure and constructs a (dis)assembly tree, which guides the target locations to move to an expanded configuration from root to leaves in Stage III.
At each level, the expanded locations serve as guideposts for the robots to follow in the subsequent stages.
Notably, the time consumption of Stages I, II and III is negligible, as they involve purely computational processes. 
In Stage III, the robots efficiently reach their respective targets in the expanded configuration, then independently trace the recorded guideposts in a decentralized manner, finally assembling into the desired structure. 
The allocation result from Stage I guarantees the final structure's connectivity, ensuring that each robot properly reaches its assigned target location.

\subsection{Stage I: Target Dispatching}

\begin{algorithm}
	\caption{Dispatching($\mathbf{G}$, $\mathbf{A}$)}\label{alg:dispatch}
	\KwIn{A set of target locations $\mathbf{G}$, a set of robots $\mathbf{A}$ with docking system layouts $\delta_{a_i}$}
	\KwOut{The dispatched robots $\mathbf{A}_g$, their orientations $\psi_{a_i}$, and the robot-target matching $T\left(\mathbf{A}_g\right)$}
	$TabuList, SoluList \leftarrow \varnothing$ ; \\
	$\bm{S} \leftarrow$ RandomInitialization($\mathbf{G},\mathbf{A}$) ; \\
	$K, T_{max}, endTime \leftarrow$ Initialization() ; \\
	$timeList \leftarrow endTime$ ; \\
	\While{$time < endTime$} {
		$Nbrs \leftarrow$ Neighbor($\bm{S}, TabuList, \mathbf{G},\mathbf{A}$) ; \\
		$T\left(\mathbf{A}_g\right), C_{lm} \leftarrow$ SearchBest($Nbrs$) ; \\
		$TabuList \leftarrow$ UpdateTabu($TabuList, T\left(\mathbf{A}_g\right)$) ; \\
		\If{$C_{lm} == C_{min}$}{
			$SoluList \leftarrow \bm{S}$ ; \\		
		}
		$time =$ TimeCounter() ; \\
		$timeList \leftarrow time$ ; \\
		$endTime \leftarrow$ Eq. \ref{eq:endtime}\\
	}
	\If{\rm{Length}($SoluList$) $> 1$}{
		$SoluList \leftarrow$ MinDistance($SoluList, \mathbf{G}$) ; \\
	}
	\If{\rm{Length}($SoluList$) $> 1$}{
		$SoluList \leftarrow$ MinExtension($SoluList$) ; \\
	}
	\If{\rm{Length}($SoluList$) $> 1$}{
		$SoluList \leftarrow$ MaxConnection($SoluList$) ; \\
	}
	\If{\rm{Length}($SoluList$) $> 1$}{
		$SoluList \leftarrow$ PickOne($SoluList$) ; \\
	}
	$\mathbf{A}_g, \psi_{a_i}, T\left(\mathbf{A}_g\right) \leftarrow SoluList$ \\
\end{algorithm}





In the first stage, all target locations are efficiently assigned to the participating robots, where the idle robots, as $N \ge M$, do not partake in the self-assembly process.
The dispatching task is formulated as an optimization problem, where the state variable $\bm{S} = \left( \mathbf{A}_g, T\left(\mathbf{A}_g\right) \right)$ represents the constructed structure involving the dispatched robots with diverse DMLs and orientations, and the matching between robots and targets.
The objective is to optimize the cost function $\mathcal{C}$, which corresponds to the connectivity number achieved when robots are placed on all target locations and connected through their docking mechanisms. 
Then, the problem can be expressed as
\begin{equation}
\begin{aligned} 
&\underset{\bm{S}}{\min}\,\,\left| \bm{S} \right|_c
\\
\text{s}.\text{t}. \quad & \left| \mathbf{A}_g \right|=M, \mathbf{A}_g\subseteq \mathbf{A},
\\
&\left| T\left(\mathbf{A}_g\right) \right|=M, \psi_{a_i}\left(t\right) \in \mathbf{\Phi}.
\end{aligned} 
\end{equation}

To identify the feasible target allocation for robots with diverse DMLs, we utilize a tabu search-based iterative algorithm, as outlined in Alg. \ref{alg:dispatch}. 
This proposed tabu search algorithm incorporates three neighborhood moves: exchanging two matches, rotating a robot, and replacing a robot. 
Importantly, since the idle robots remain unconnected, we can establish the global minimal cost $\mathcal{C}_{min}$ as
\begin{equation}
\mathcal{C}_{min} = N - M + 1 .
\end{equation}

In each iteration, the algorithm efficiently identifies the best robot-target mapping, along with the appropriate robot orientation $T\left(\mathbf{A}_g\right)$, by exploring the local minimal cost $\mathcal{C}_{lm}$ within the neighborhood of the last mapping. 
The time complexity of searching the neighborhood in line 6 and 7 is $O \left( N^2 \right)$ and updating the tabu list in line 8 has a time complexity of $O \left( N \cdot L_{tabu} \right)$, where $L_{tabu}$ is the length of the tabu list. 
If $\mathcal{C}_{lm}$ is equal to $\mathcal{C}_{min}$, the current solution is added to the candidate list.
This search process continues until it either takes too long to find the next solution or reaches the preset maximum number of loops.
To avoid the algorithm getting stuck while seeking a few and scattered residual solutions, we estimate the upper bound of the consumed time $t_i$ in the $i$-th iteration, denoted as $endTime$ in Alg. \ref{alg:dispatch}. This estimation is achieved by amplifying the average time of all past iterations based on the moving average method, given by
\begin{equation} \label{eq:endtime}
t_{i}=K\frac{\sum_{j=1}^{i-1}{t_j}+T_0}{i},
\end{equation}
where $K$ and $T_0$ are the user-defined coefficients. 

Next, three sequential criteria are employed to select the optimal solution from the previously generated candidate solutions. These criteria include the shortest total path from robots to targets, the most balanced self-assembly tree, and the maximum connections between robots.
In line 15, Alg. \ref{alg:dispatch} utilizes criterion 1 to choose solutions with the shortest total path from robots to targets, optimizing the time required for movement.
In line 17, criterion 2 is applied to select solutions with the most balanced self-assembly trees, maximizing the efficiency of the docking process.
Furthermore, in line 19, criterion 3 is used to select solutions with the maximum connections between robots, ensuring a stronger and more interconnected final structure.
If multiple optimal solutions are identified through the above criteria, any one of them can be adopted for further steps.


\subsection{Assembly Tree Generation}

\begin{algorithm}
	\caption{TreeGeneration($G$)}\label{alg:assemblytree}
	\KwIn{all final targets $G$, location of the target extension center $c$, the seperating distance of one exntension step $d$}
	\KwOut{Assembly tree $Tree$}
	
	\SetKw{Return}{return}
	\SetKw{Break}{break}
	\SetKw{And}{and}
	
	/*When group $G$ contains one point, ends*/ \\
	\If{$|G| == 1$}{\Return}
	/*Find all possible partitions by lines.*/\\
	$P \leftarrow$ AllDivisions($G$);\\
	/*Solve Eq. \ref{eq:division} for the best division $G_1,G_2$.*/\\
	\ForEach{($G_i, G_j \in P$)}{
		\If{$\lvert G_i\rvert_c \lvert G_j \rvert_c == 1$}{
			$(G_1,G_2) \leftarrow $  BestDivision($G_i, G_j$);\\
		}
	}
	/*Create new node to save the two $G_1,G_2$.*/\\
	$G.lChild \leftarrow $ NewNode($G_1$);\\
	$G.rChild \leftarrow $ NewNode($G_2$);\\
	TreeGeneration($G.lChild$);\\
	TreeGeneration($G.rChild$);\\
\end{algorithm}

In this stage, the generation of a binary assembly tree is accomplished through a recursive algorithm, performed in line 17 of Alg. \ref{alg:dispatch}, to guide the (dis)assembly process. 
This algorithm is modified from \cite{saldana2017decentralized}, with the key difference being that, at each recursion, both split parts of the assembly tree are retained as connected structures.

Alg. \ref{alg:assemblytree} outlines the process of dividing the target locations set $G$ into two groups $G_i$ and $G_j$ using a horizontal or vertical line, ensuring that each group remains internally connected.
This division is followed by recursively splitting $G_i$ and $G_j$ into two subgroups, while maintaining internal connectivity in each subgroup. 
This recursive process continues until only one point remains in each group.
To achieve a balanced division at each recursion, an optimization problem is solved,
\begin{equation} \label{eq:division}
\begin{aligned} 
&\underset{G_i,G_j}{\max} \; f\left( G_i,G_j \right) 
\\
\text{s}.\text{t}. \quad &\left| G_i \right|+\left| G_j \right|=\left| G \right|,
\\
 &\lvert G_i\rvert_c \lvert G_j \rvert_c = 1,
\end{aligned} 
\end{equation}
where $f\left( G_i,G_j \right)=\left| G_i \right|\left| G_j \right|$ is a factor to evaluate each division. 
This algorithm is modified from the Alg. 1 in \cite{saldana2017decentralized} without changing complexity, so the time complexity is $O(M^3 \log M)$.

\subsection{Target Extension}

\begin{algorithm}
	\caption{ExtendTarget($Tree$, $G$, $c$)}\label{alg:extend}
	\KwIn{Assembly tree $Tree$, target group to be separated $G$, extension center $c$}
	\KwOut{A series of expanded target locations $E$}
	
	\SetKw{Static}{static}
	\SetKw{Return}{return}
	\SetKw{Break}{break}
	\SetKw{And}{and}
	
	/*End until group $G$ contains one point*/ \\
	\If{$|G| == 1$}{\Return}
	
	/*$G_m$/$G_u$ is the moving/unmoving group.*/ \\
	$G_m,G_u,dir \leftarrow$ Split($Tree, G, c$) ; \\
	\If{$dir==$`$x$'}{
		$G_m.x \leftarrow $ Eq. \ref{eq:extension}a ; \\
	}
	\ElseIf{$dir==$`$y$'}{
		$G_m.y \leftarrow $ Eq. \ref{eq:extension}b ; \\	
	}
	$G_m$.Extend(); \\
	$G_m.c\leftarrow $ ClosestPoint($G_m$, $c$) ; \\
	
	ExtendTarget($Tree$, $G_m$, $G_m.c$) ; \\
	ExtendTarget($Tree$, $G_u$, $c$) ; \\
	
\end{algorithm}


To ensure a smooth and gradual extension of the target structure, Alg. \ref{alg:extend} introduces a progressive approach based on the generated binary assembly tree. Unlike the one-step mapping method used in \cite{saldana2017decentralized}, which skips intermediate docking positions, the proposed algorithm expands the desired structure in a top-down level order of the assembly tree.
The moving distance $L_e$ of the expanded group is
\begin{equation}
L_e=I_e\left( W_e+1 \right) -1,
\end{equation}
where $I_e$ denotes the interval between two targets after complete extension, and $W_e$ is the width in the separation direction of the to-be-split structure.

Based on the direction indicated by the assembly tree, Alg. \ref{alg:extend} calculates the new locations for two groups in the same pair that are to be separated during each expansion step. The new positions are determined using the following equations:
\begin{subequations} \label{eq:extension}
\begin{align} 
x &= x+sign\left( x-x_c \right) L_e,
\\
y &= y+sign\left( y-y_c \right) L_e,
\end{align} 
\end{subequations}
where $x$ and $y$ denote the locations of the robot group, while $x_c$ and $y_c$ are the location of the extension center of the target group that is to be split.
Lastly, a new center will be assigned to the point that is closest to the center in the previous moving group. 
This process continues iteratively until there is only one point contained in the group, which signifies the full extension for all targets. 

In each level of the assembly tree, all target points are visited or updated. The assembly tree generated from a structure with $M$ locations has at most $M-1$ levels. Therefore, the overall access number could be $M\left( M-1 \right)$. As a result, the time complexity of Alg. \ref{alg:extend} is $O \left( M^2 \right)$. 

\subsection{Robot Navigation}
\label{sect:navigation}

\begin{algorithm} [h]
	\caption{Navigation($Tree$, $Map$, $E$, $R$)}
	\label{alg:robotnavigation}
	\KwIn{assembly tree $Tree$, map with obstacles $Map$, a set of landmarks $E$, all robots $R$}
	\KwOut{desired strucuture built by robots}
	/*Initialize robots from $Tree$ leaves.*/ \\
	\ForEach{$R_i \in R$}{
		$R_i$.GetTarget($Tree$) ; \\
		$R_i$.GetOrientation($Tree$) ; \\
		$R_i$.Rotate(); \\
	}
	\ForEach{ \emph{level} l \emph{of} $Tree$, \emph{from leaves to root}}{
		
		$G \leftarrow \varnothing$; /*To save all robot groups.*/ \\
		\ForEach{Node \emph{in level} l}{
			/*Initialize each robot group.*/\\
			$G_i \leftarrow$ NewGroup($Node$, $R$); \\
			$E_i \leftarrow$ GetLandmark($G_i$, $E$); \\
			/*Plan paths from local information.*/ \\
			$G_i$.PerceptionAndPathPlanning($Map$, $E_i$); \\
			Save2G($G_i$, $G$); /*Save $G_i$ to $G$.*/\\
		}
		
		\While{not all $G$ \emph{reached targets}}{
			\ForEach{$G_i \in G$}{
				$G_i$.MoveAlongPath(); \\
				\If{\emph{$G_j$ in $G_i$ partner's target region}}{
					/*Authorized to adjoin $G_j$.*/ \\
					$G_i$.AllowCloseToPartner($G_j$); \\
					$G_i$.TopPriority(); \\
					$G_i$.Move2DockPartner($Map$); \\
				}
				/*Re-plan if stuck.*/ \\
				\If{$G_i$\emph{'s move blocked}}{
					$G_i$.PathPlanning($Map$, $E_i$); \\
				}
			}
			/*dock once adjacent.*/ \\
			\ForEach{$G_i, G_j \in G$}{
				\If{$G_i$ and $G_j$ adjacent}{
					$G_k \leftarrow$ Dock($G_i$, $G_j$); \\
				}
			}
			\If{loop infinite}{
				return $Fail$; \\
			}
		}
	}
\end{algorithm}

In this stage, the robots with allocated targets plan their trajectories in $O(k \log k)$ time ($k$ is the number of map grids) and move in a distributed manner \cite{Zhang2023Parallel}, as shown in Alg. \ref{alg:robotnavigation}.
Initially, robots rotate into their desired directions and then navigate themselves following the bottom-up order of the assembly tree.
During robot movement, all robot groups maintain a distance of at least 1 cell from each other, and collision avoidance rules are followed, as depicted below in Fig. \ref{fig:collisionAvoid}.
\emph{Rule for collision avoidance}: 
\begin{enumerate}
\item If a group $G_i$ is stopped by another group $G_j$ during movement, $G_i$ will maneuver around $G_j$ by involving it in the path re-planning, while $G_j$ will continue moving.
\item If group $G_i$ and $G_j$ mutually block each other, with $G_i$ owning a lower predetermined priority, $G_i$ will navigate around $G_j$ by re-planning the path, while $G_j$ will wait for a preset number of steps before continuing.
\end{enumerate}

\begin{figure} [htbp]  
	\centering
	\subfloat[]{\label{fig:collisionAvoidi}         
		\begin{minipage}[b]{0.4\linewidth}      
			\centering      
			\includegraphics[width=1\linewidth]{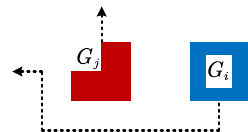}      
		\end{minipage} 
	}
	\subfloat[]{\label{fig:collisionAvoidii}
		\begin{minipage}[b]{0.4\linewidth}      
			\centering      
			\includegraphics[width=1\linewidth]{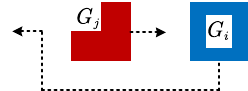}      
		\end{minipage}      
	} 
	\caption{Rule (1) in panel (a) and (2) in panel (b) for collision avoidance.}
	\label{fig:collisionAvoid}
\end{figure}

When a group $G_i$ perceives that another group $G_j$ has completely occupied its partner's target region, $G_i$ will initiate the docking action. Instantly, the priority of $G_i$ will be raised to the highest level to avoid any potential interruption from other groups. Once they adjoin each other, $G_i$ directly moves towards and docks with $G_j$, ensuring a seamless and efficient assembly process.

\newcommand{\gridmapwidth}{0.5}
\begin{figure} [htbp]
	\centering
	\subfloat[]{\label{fig:simuMaps0}         
		\begin{minipage}{\gridmapwidth\linewidth}      
			\centering      
			\includegraphics[width=1\linewidth]{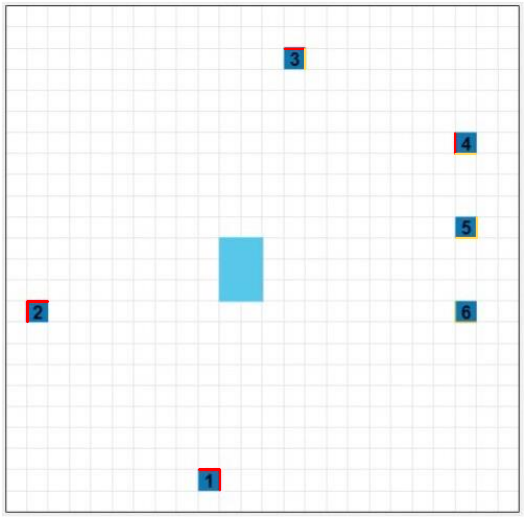}      
		\end{minipage} 
	}\subfloat[]{\label{fig:simuMaps1}
		\begin{minipage}{\gridmapwidth\linewidth}      
			\centering           
			\includegraphics[width=1\linewidth]{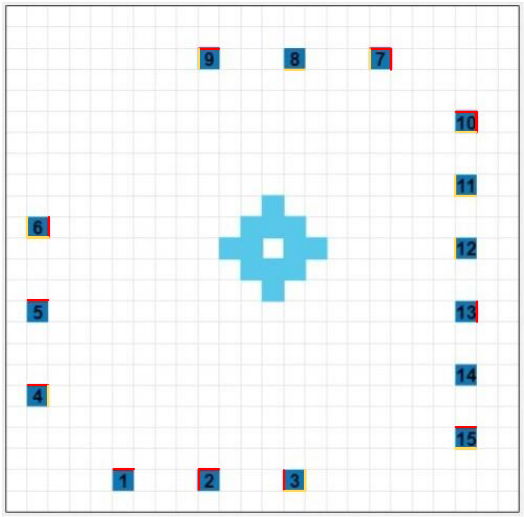}      
		\end{minipage}      
	}
	
	\subfloat[]{\label{fig:simuMaps2}
		\begin{minipage}{\gridmapwidth\linewidth}      
			\centering            
			\includegraphics[width=1\linewidth]{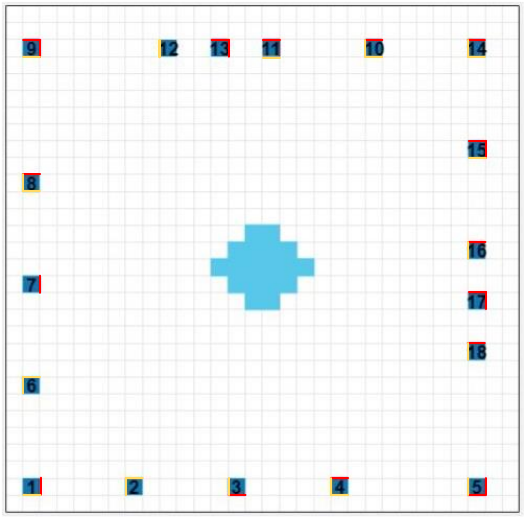}      
		\end{minipage}      
	}\subfloat[]{\label{fig:simuMaps3}
		\begin{minipage}{\gridmapwidth\linewidth}      
			\centering           
			\includegraphics[width=1\linewidth]{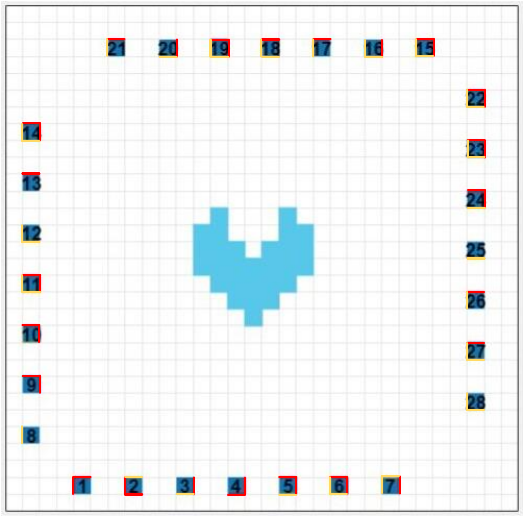}      
		\end{minipage}      
	}   
	\caption{Four simulation maps. Cells in blue and cerulean denote robots and targets, respectively. The red and yellow edges represent the gendered docking systems, while the genderless ones will be denoted by monochromatic edges.}
	\label{fig:simuMaps}
\end{figure}

\section{SIMULATION EVALUATION}
\label{sect:evaluation}

\subsection{Simulation Configuration} 
\label{sect:simulationConfiguration}

To validate the effectiveness and efficiency of our algorithm, we have implemented it in MATLAB as an open-source project available at https://github.com/Joyyy821/DockPlanning. 
During the simulations, we utilized four different grid maps, each representing distinct target structures, as illustrated in the Fig. \ref{fig:simuMaps}.
For each map, the algorithm was executed 20 times to obtain statistical averages of feasile solutions for robot-target matching and the number of robot moving steps. 
The maps gradually increased in scale and complexity from the first to the fourth, posing greater computational challenges.



As depicted in Fig. \ref{fig:cuboat}, CuBoats can be outfitted with either genderless or gender-opposite docking mechanisms, abbreviated as Genderless or Gender-opposite. In both scenarios, the DMLs of all robots vary.
In the following sections, we will compare the performance of our SAP algorithm in these scenarios, highlighting its adaptability and robustness in different robotic settings.

\subsection{Simulation Results} 
\label{sect:simulationResults}

\begin{table*}[]
	\centering
	\caption{SIMULATION RESULTS}
	\label{tab:simulation}
	\begin{tabular}{|c|c|c|c|cc|cc|cc|cc|}
		\hline
		\multicolumn{1}{|c|}{\multirow{2}{*}{Map}} & \multicolumn{1}{c|}{\multirow{2}{*}{Targets $M$}} & \multicolumn{1}{c|}{\multirow{2}{*}{Robots $N$}} & \multicolumn{1}{c|}{\multirow{2}{*}{Docking Systems}} & \multicolumn{2}{c|}{Average Solutions}                                 & \multicolumn{2}{c|}{STD of Solutions}                                  & \multicolumn{2}{c|}{Average Steps}                                     & \multicolumn{2}{c|}{STD of Steps}                                      \\ \cline{5-12} 
		\multicolumn{1}{|c|}{}                     & \multicolumn{1}{c|}{}                         & \multicolumn{1}{c|}{}                        & \multicolumn{1}{c|}{}                                  & \multicolumn{1}{c|}{G-less} & \multicolumn{1}{c|}{G-opposite} & \multicolumn{1}{c|}{G-less} & \multicolumn{1}{c|}{G-opposite} & \multicolumn{1}{c|}{G-less} & \multicolumn{1}{c|}{G-opposite} & \multicolumn{1}{c|}{G-less} & \multicolumn{1}{c|}{G-opposite} \\ \hline
		1   &  6  & 6 &  12  & \multicolumn{1}{c|}{6.40}         &  \textbf{8.53}  & \multicolumn{1}{l|}{0.66}   &  2.98   & \multicolumn{1}{l|}{\textbf{28.37}}    &  {29.57}  & \multicolumn{1}{l|}{1.47}    &   2.50      \\ \hline
		2    &  12  & 15 &  27  & \multicolumn{1}{c|}{\textbf{15.31}}         &  7.09  & \multicolumn{1}{l|}{7.94}   &  4.03   & \multicolumn{1}{l|}{{53.89}}    &  \textbf{49}  & \multicolumn{1}{l|}{13.99}    &   5.25      \\ \hline
		3   &  18  & 18 &  42  & \multicolumn{1}{c|}{\textbf{182.31}}         &  46.33  & \multicolumn{1}{l|}{48.75}   &  24.81   & \multicolumn{1}{l|}{{64.05}}    &  \textbf{62.78}  & \multicolumn{1}{l|}{6.26}    &   7.33      \\ \hline
		4   &  28  & 28 &  78  & \multicolumn{1}{c|}{\textbf{408.82}}         &  259.10  & \multicolumn{1}{l|}{177.73}   &  131.92   & \multicolumn{1}{l|}{{82.19}}    &  \textbf{78}  & \multicolumn{1}{l|}{10.66}    &   10.67      \\ \hline
		\multicolumn{12}{l}{\scriptsize $^*$ G-less: Genderless, G-opposite: Gender-opposite, STD: Standard Deviation.}
	\end{tabular}
\end{table*}

\begin{figure}[htbp]      
	\centering      	
	\subfloat[]{\label{fig:solutions}       
	\begin{minipage}{0.85\linewidth}      
		\centering      
		\includegraphics[width=1\linewidth]{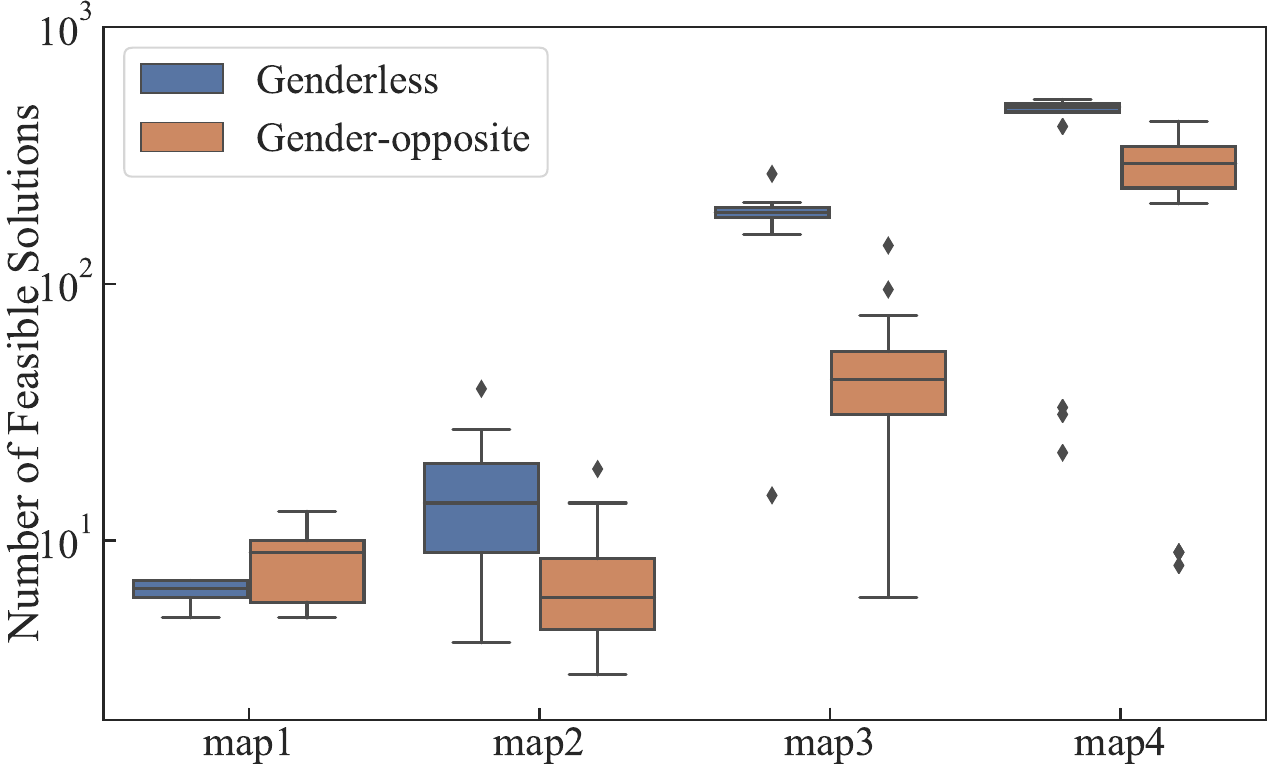}
	\end{minipage}      
	} 

	\subfloat[]{\label{fig:robotStep}       
		\begin{minipage}{0.85\linewidth}      
			\centering      
			\includegraphics[width=1\linewidth]{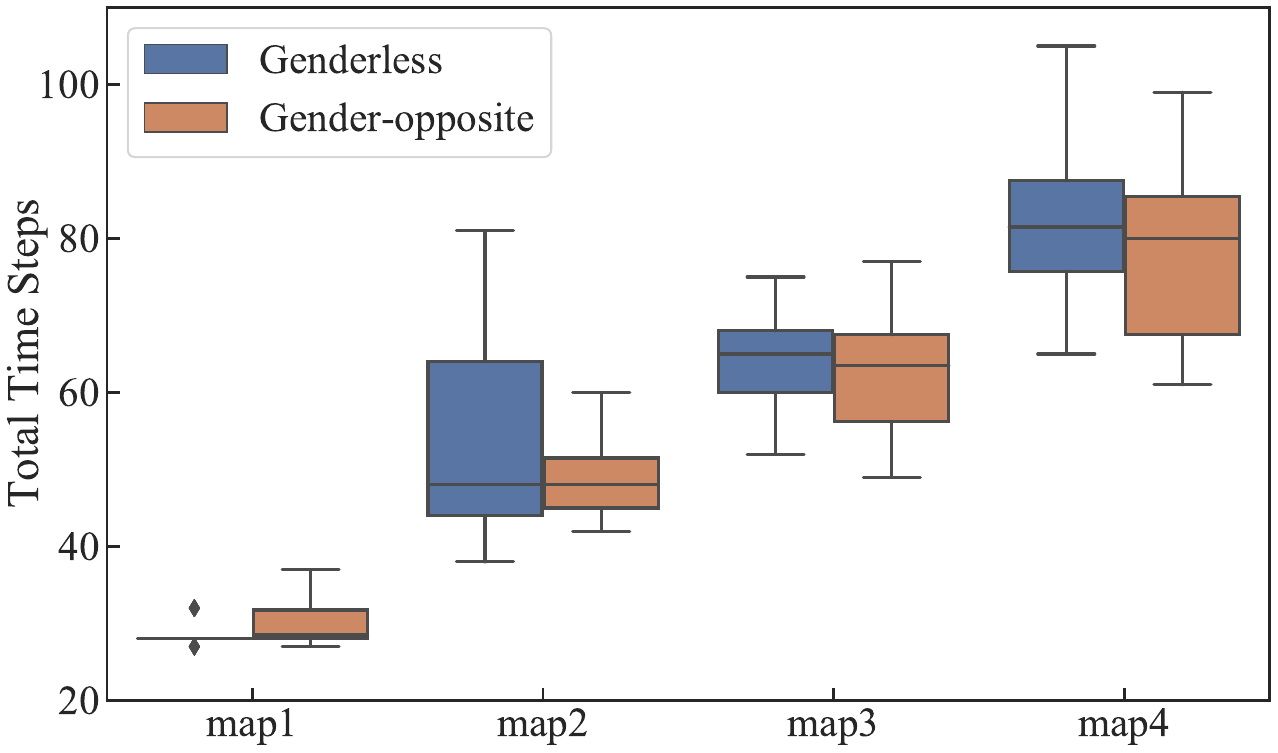}      
		\end{minipage}      
	}      
	\caption{Simulations of the SAP algorithm in the Genderless and Gender-opposite scenarios across 4 designed maps for comparison. (a) The number of feasible solutions for robot-target matching.  (b) The robot moving steps for assembly.}   
	\label{fig:simulationResults}  
\end{figure}

The simulation results presented in Fig. \ref{fig:simulationResults} and Table \ref{tab:simulation} indicate several important trends in the performance of our SAP algorithm under the Genderless and Gender-opposite scenarios.
Firstly, as the number of robots and target points increases, there is a noticeable ascend in the number of feasible solutions for target allocation and the robot moving steps. This observation suggests that as complexity increases, exploring a larger solution space poses greater efforts for self-assembly.
Fig. \ref{fig:solutions} reveals that in most maps, the algorithm can find more feasible solutions for assembly in the Genderless scenario than that in the Gender-opposite one. 
This is because the former imposes fewer constraints on forming the target structures.
Lastly, Fig. \ref{fig:robotStep} demonstrates that the Gender-opposite scenario results in the fewer robot moving steps, indicating that the algorithm in this scenario is more efficient in terms of optimizing the self-assembly process in Alg. \ref{alg:dispatch}. 
In contrast, the Genderless scenario, despite having more feasible solutions, exhibits more robot moving steps, suggesting lower efficiency. 
This finding undermines the importance of docking system gender in determining the efficiency of self-assembly processes.


\begin{figure} [tbp] 
	\centering
	\includegraphics[width=0.8\linewidth]{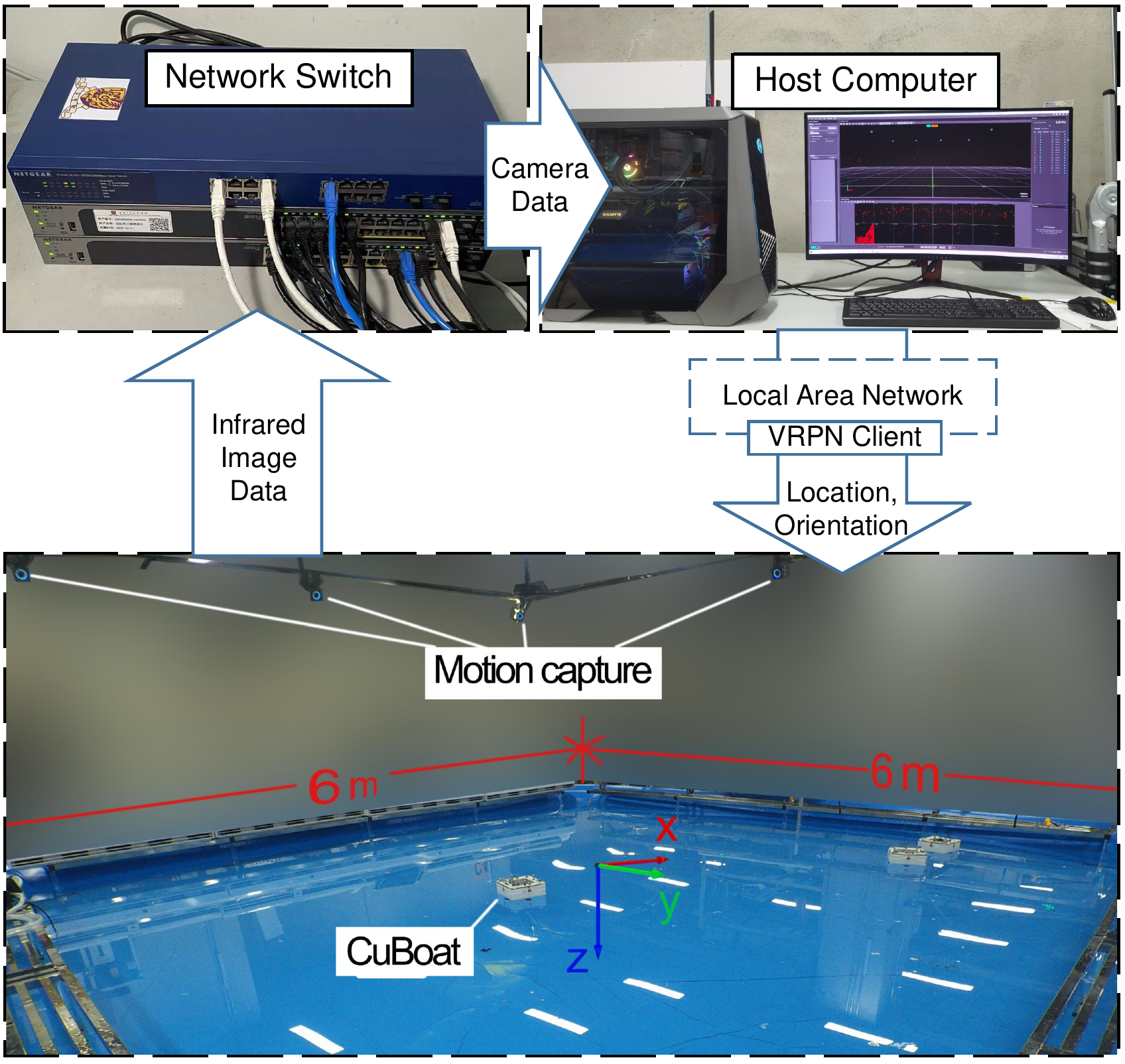}
	\caption{Experimental setup. }
	\vspace{-10pt}
	\label{fig:dock_scene}
\end{figure}

\section{Experiments}
\label{sect:experiment}

\subsection{Experimental Setup}
This section encompasses two experiments carried out in a $6 m \times 6 m$ pool to validate the tracking performance of the controller and demonstrate the practical deployment of the SAP algorithm. 
The experimental setup is exhibited in Fig. \ref{fig:dock_scene}.
To facilitate tracking, infrared cameras are utilized to capture image data, which is then processed by the motion capture system to calculate the global positions and orientations of all CuBoats. 
As aforementioned, each CuBoat receives its localization data through a VRPN interface, enabling decentralized navigation and movement coordination during the experiments.

\begin{figure} [tbp] 
	\centering
	\includegraphics[width=0.8\linewidth]{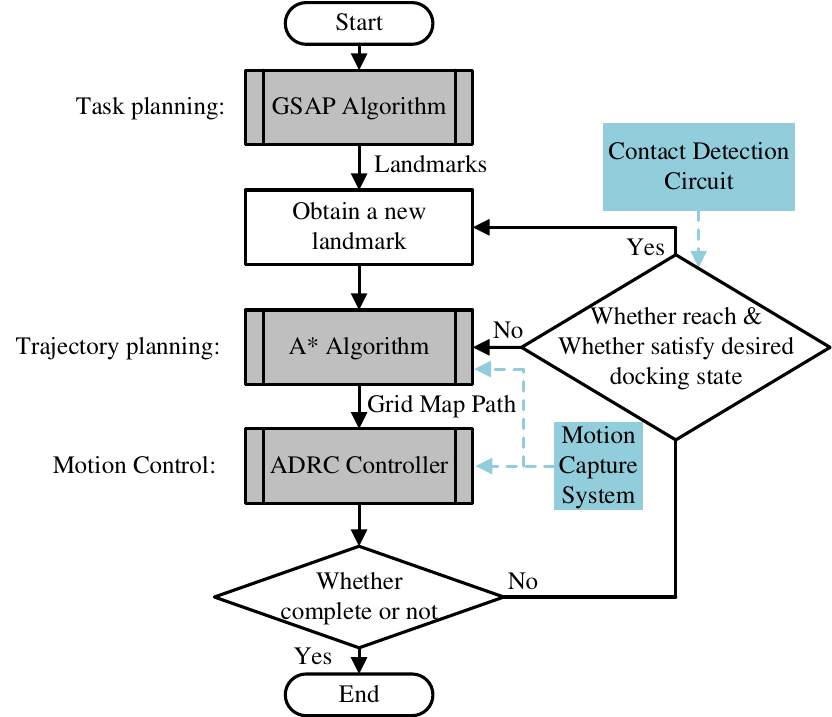}
	\caption{Architecture of the navigation system. }
	\vspace{-10pt}
	\label{fig:control_frame}
\end{figure}

Fig. \ref{fig:control_frame} illustrates how the CuBoats can autonomously navigate using the deployed SAP algorithm.
Initially, the SAP algorithm operates on a central computer, generating guidepost points that are subsequently transmitted to each dispatched CuBoat. 
Subsequently, the CuBoats individually select these guideposts as their current targets and perform real-time trajectory planning using the A* algorithm, as described in Section \ref{sect:navigation}. 
Then, the CuBoats track their planned trajectories by employing the ADRC controllers.
To track their planned trajectories, the CuBoats utilize ADRC controllers as low-level controllers during the experiments.
To ensure smooth movement, the pool is meshed into small cells of size $ 0.25 \text{m} \times 0.25 \text{m}$, resulting in large gaps among path points.
To prevent jerky motion, the thrust of each CuBoat's thrusters is limited within a user-defined range $ \left[ f_{min},f_{max} \right]$.
Importantly, all CuBoats can only access the position information of their neighboring CuBoats, allowing them to execute decentralized movement. 


\begin{figure*}[htbp]      
	\centering           
	\begin{minipage}{0.24\linewidth}      
		\centering
		\subfloat[]{      
			\includegraphics[width=1\linewidth]{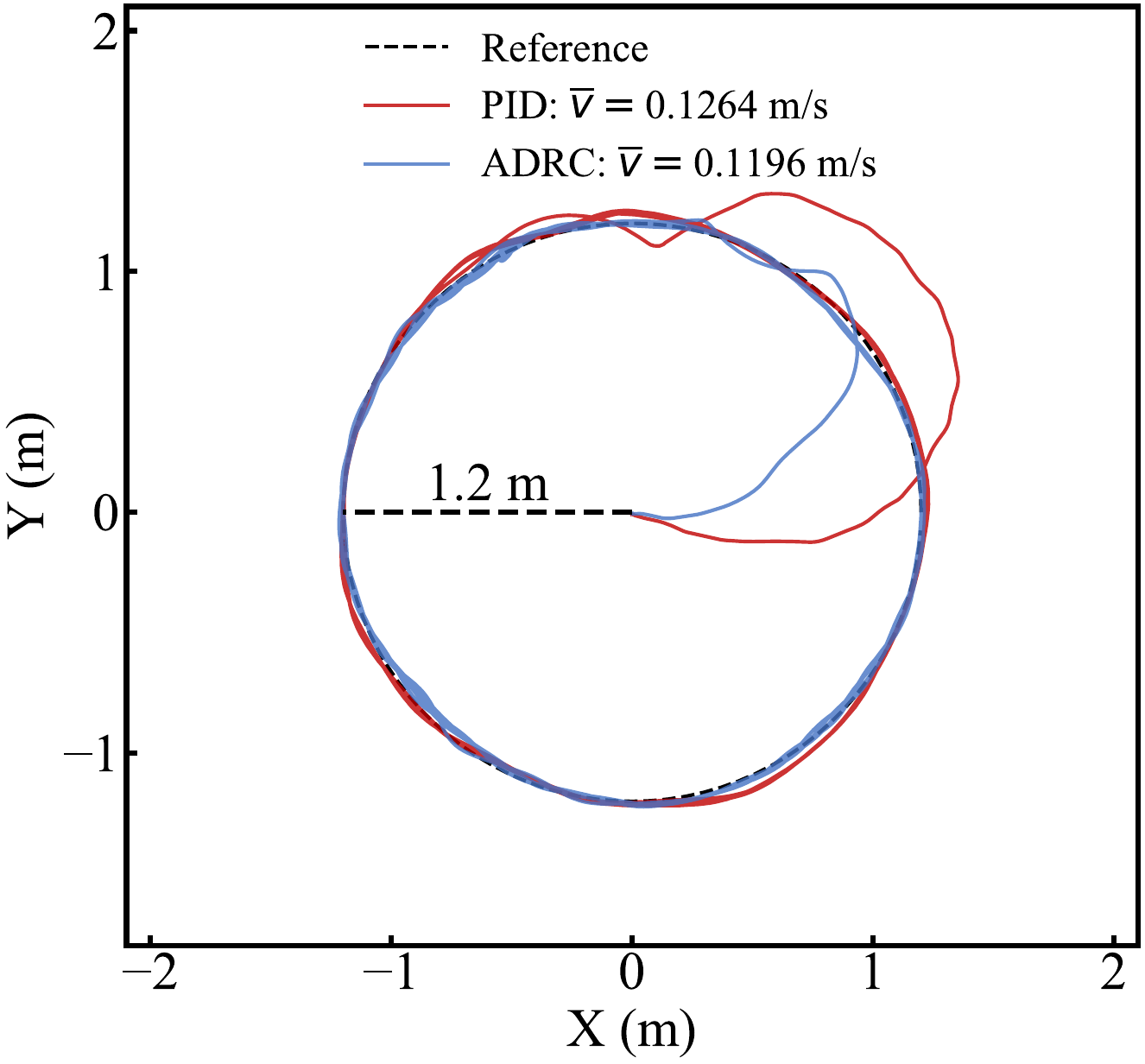}  
		}
	\end{minipage}        
	\begin{minipage}{0.24\linewidth}      
		\centering 
		\subfloat[]{         
			\includegraphics[width=1\linewidth]{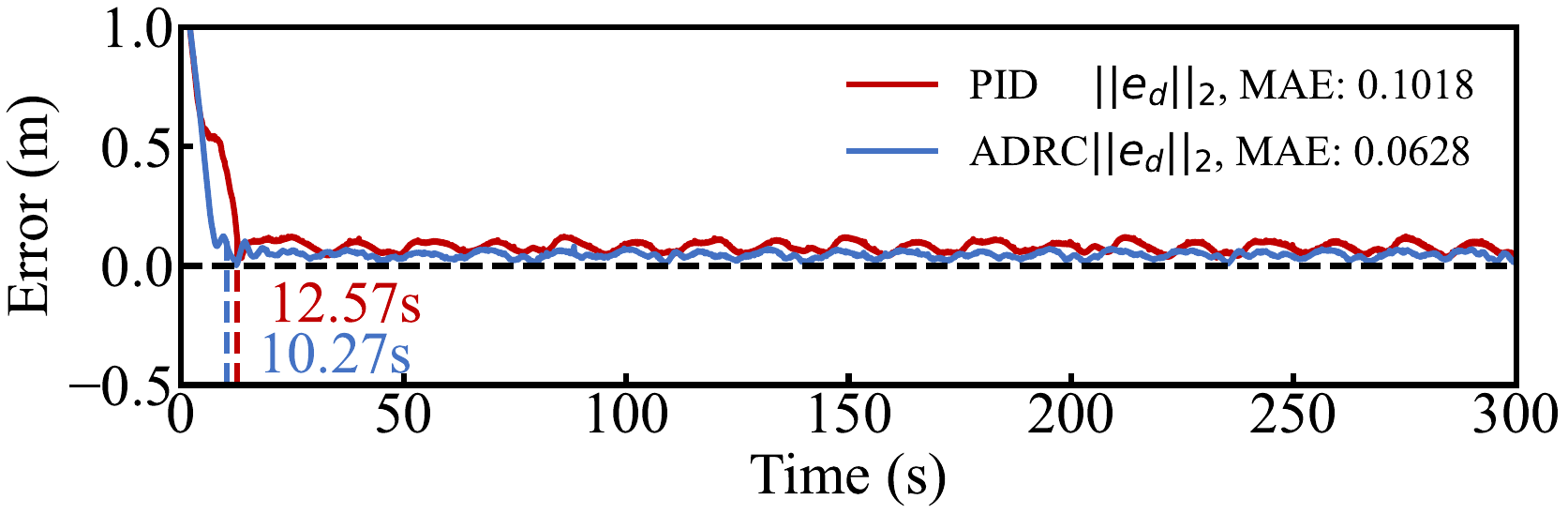} 
		}  
		\vspace{0cm}
		\subfloat[]{
			\includegraphics[width=1\textwidth]{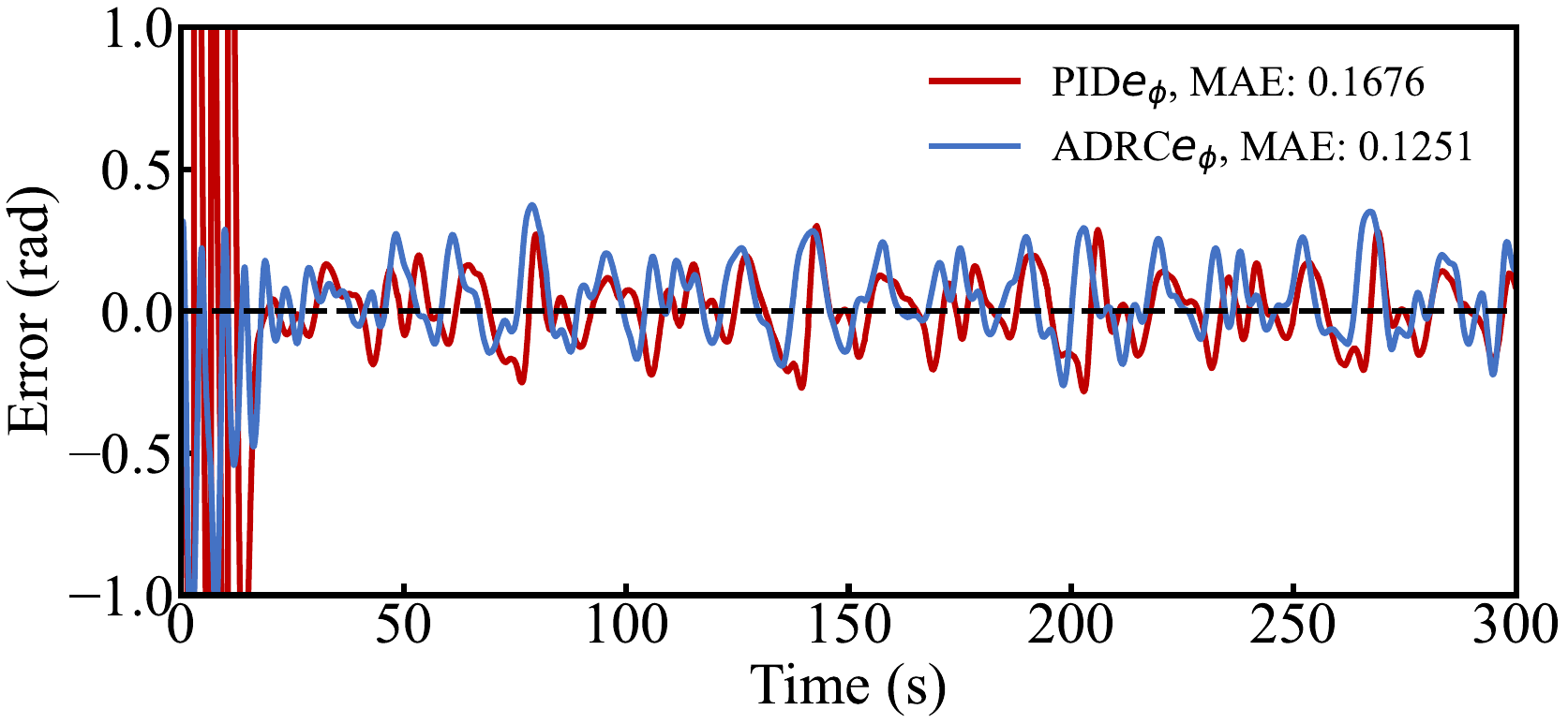}
		}
	\end{minipage}      
	\begin{minipage}{0.24\linewidth}      
		\centering
		\subfloat[]{      
			\includegraphics[width=1\linewidth]{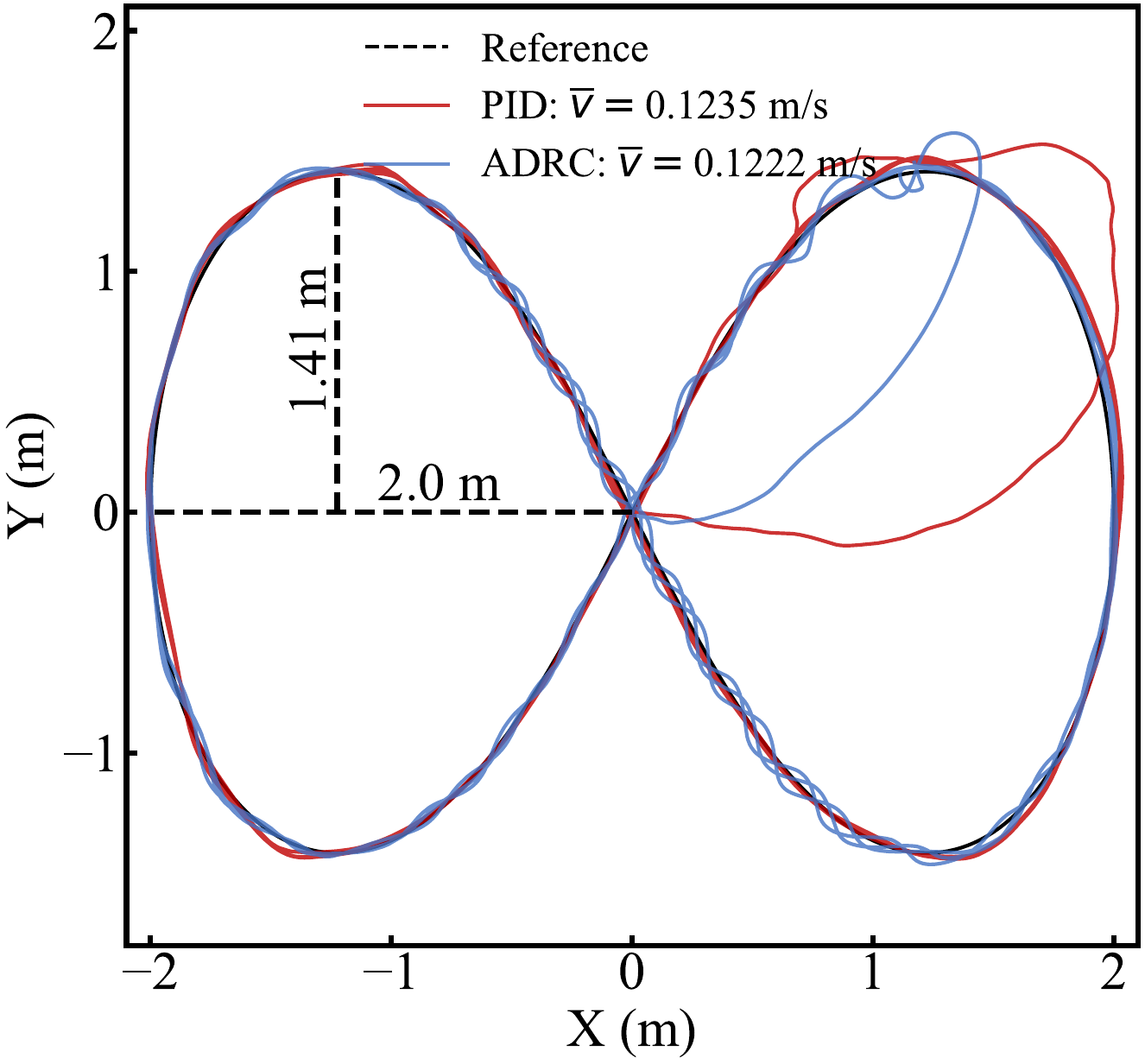}  
		}
	\end{minipage}        
	\begin{minipage}{0.24\linewidth}      
		\centering 
		\subfloat[]{         
			\includegraphics[width=1\linewidth]{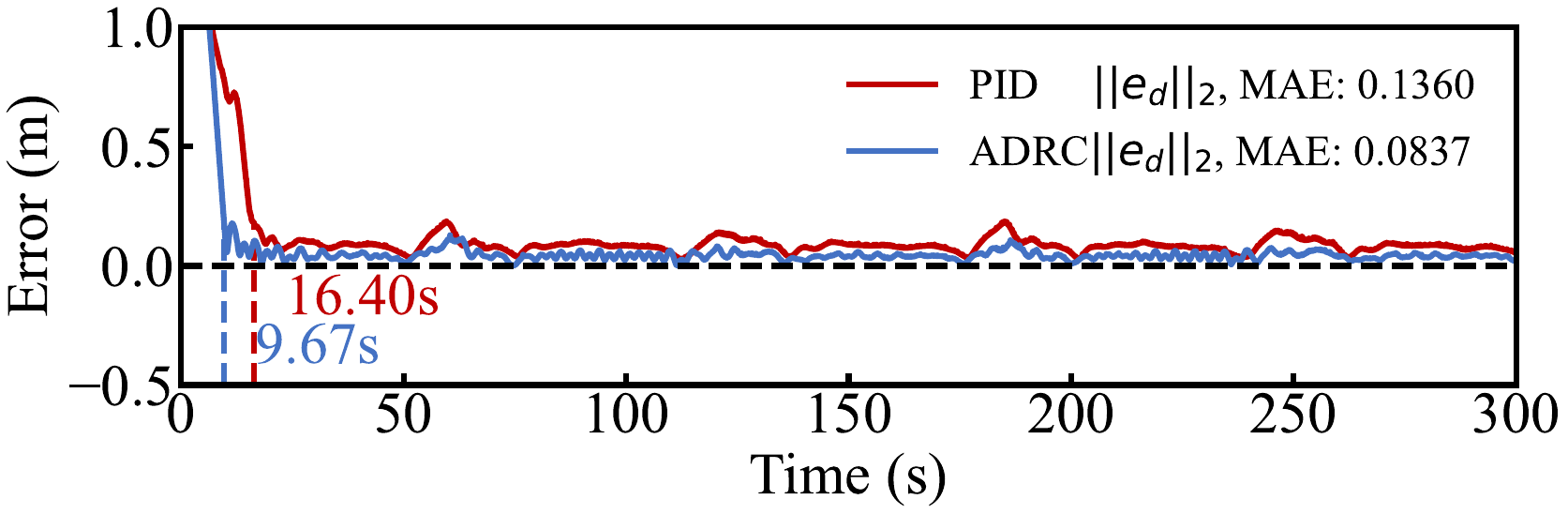} 
		}  
		\vspace{0cm}
		\subfloat[]{
			\includegraphics[width=1\textwidth]{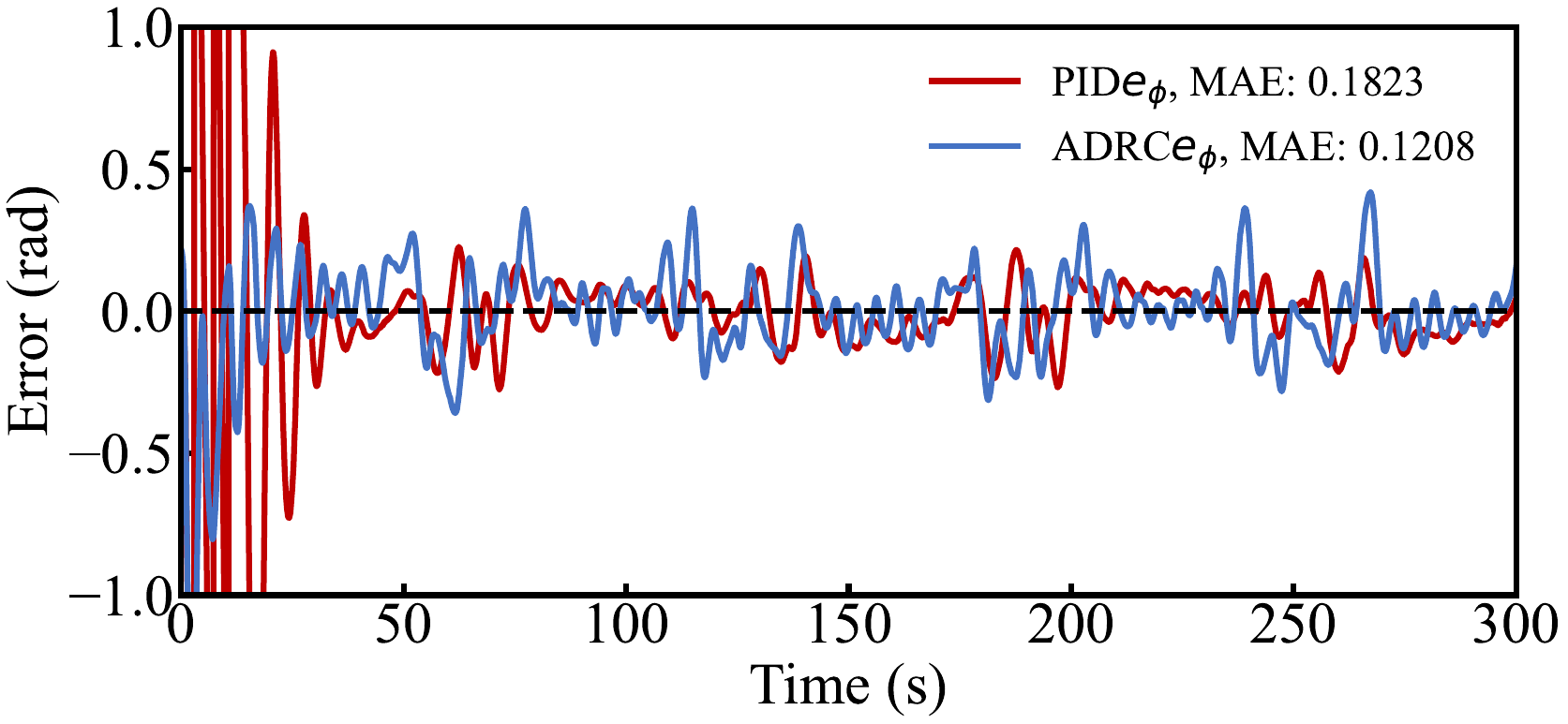}
		}
	\end{minipage}
	\caption{(a) (d) Experimental paths, (b) (e) position errors, and (c) (f) orientation errors in tracking circular and eight-pattern trajectories, respectively. The average veocities $\bar{v}$ and the mean absolute errors (MAEs) are calculated. $e_d = (e_x, e_y)$ stands for the overall error vector, where $e_x$ and $e_y$ are the errors in the X and Y directions, respectively.}   
	\label{fig:tracking}  
\end{figure*}

\subsection{Trajectory Tracking}
%
To validate the efficacy of the ADRC controller, we conducted two trajectory tracking tests: one following a circular path with a radius of $1.2$ m and the other tracing a figure-eight pattern described by $x(t) = 2\cos(t) / (1 + \sin^2(t))$ and $y(t) = 4\sin(t) \cos(t) / (1 + \sin^2(t))$. 
For each test, we applied both the fine-tuned PID and ADRC control methods. 
The PID controller gains ($K_P, K_I, K_D$) were determined using the Ziegler-Nichols method \cite{McCormack1998Rule}, as presented in \cite{Zhang2023Parallel}.
Fig. \ref{fig:tracking} portrays the experimental trajectories and the corresponding tracking errors, leading to three key observations.

The comparison between the tracking errors of the two control schemes reveals the superiority of the ADRC over the PID controller in both the circular and eight-pattern tests.
In the ADRC trials, the position errors decreased by $38.3\%$ and $38.5\%$, and the orientation errors decreased by $25.4\%$ and $33.7\%$ compared to the PID trials for the circular and eight-pattern tests, respectively. 
Moreover, the higher tracking errors observed in the eight-pattern tests, as opposed to the circular tests, confirms the increased difficulty of the figure-eight trajectory. 
Furthermore, the ADRC controller exhibits faster convergence rates with smaller overshoots in both tests, taking 10.27 seconds and 9.67 seconds for the circular and eight-pattern tests, respectively. In contrast, the PID controller takes 12.57 seconds and 16.40 seconds to achieve convergence in the corresponding tests. 
Based on these observations, we can conclude that the ADRC controller is more effective and suitable for the self-assembly tasks, providing precise and efficient navigation and control for the CuBoats.

\subsection{Self-assembly}

\newcommand{\pathfigwidth}{0.19}
\begin{figure*} [htbp]
	\centering
	\begin{minipage}{\pathfigwidth\linewidth}      
		\centering
		\subfloat[]{      
			\includegraphics[width=1\linewidth]{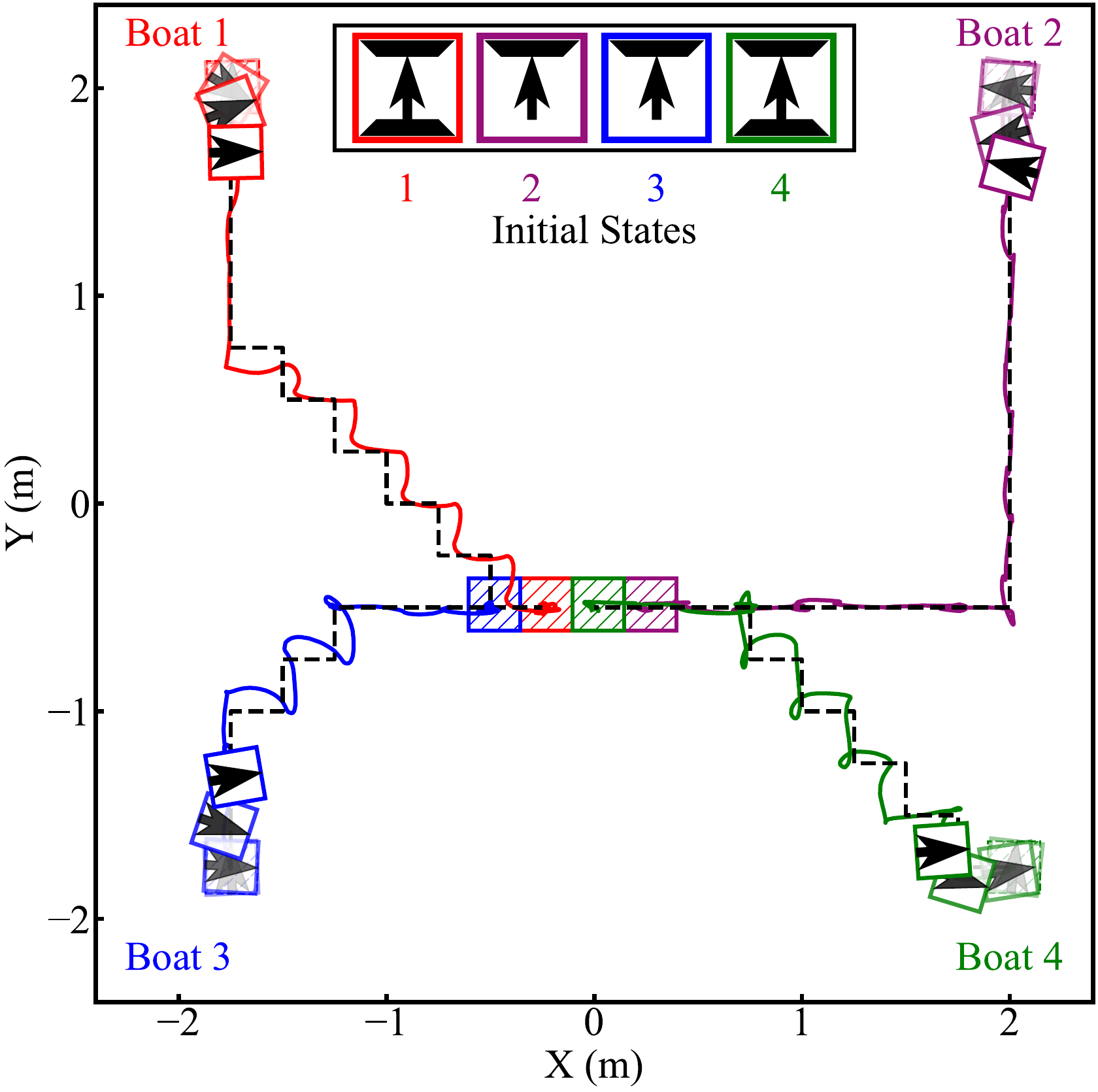}  
		}
	\end{minipage}
	\begin{minipage}{\pathfigwidth\linewidth}      
		\centering
		\subfloat[]{      
			\includegraphics[width=1\linewidth]{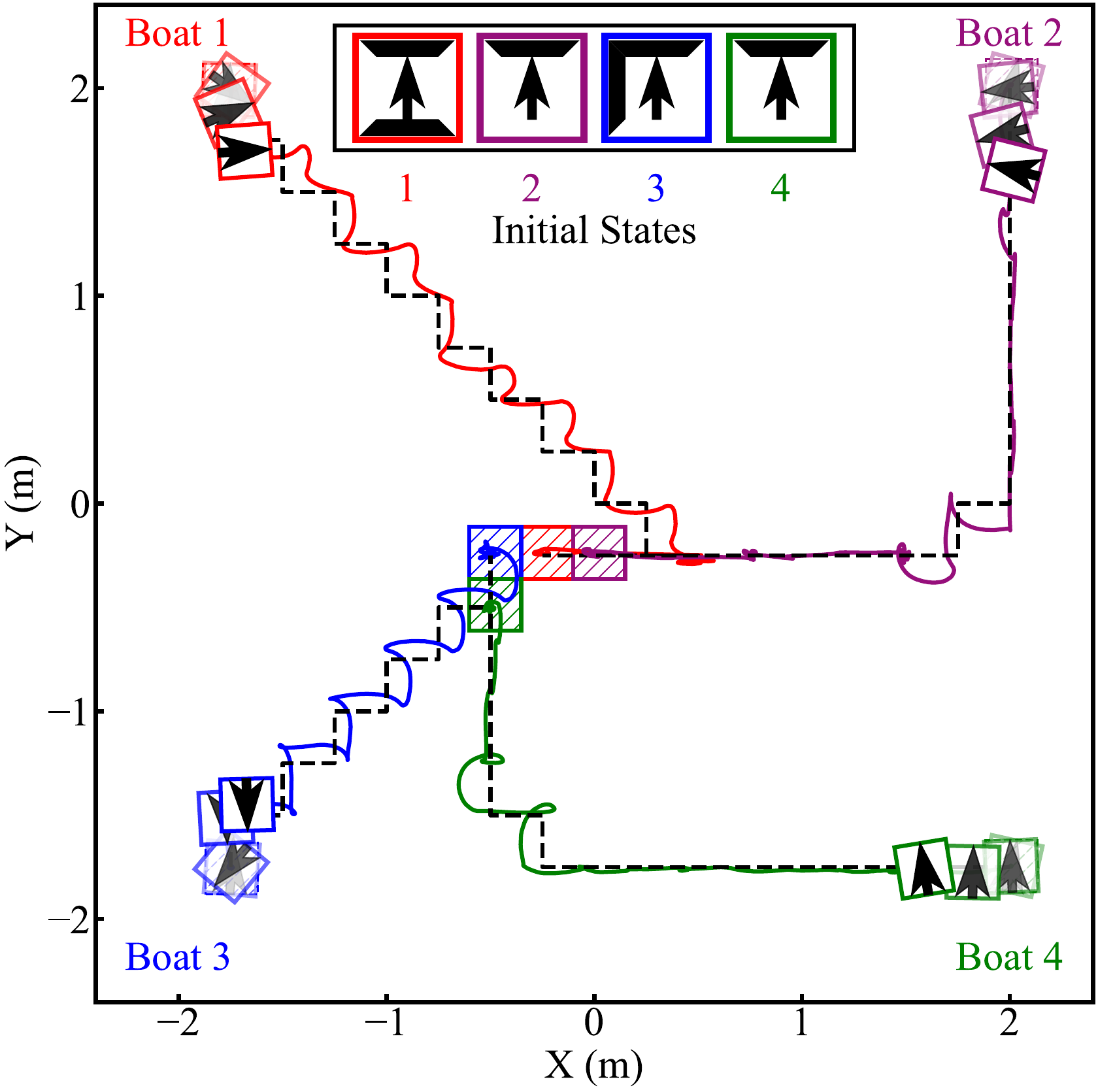}  
		}
	\end{minipage}
	\begin{minipage}{\pathfigwidth\linewidth}      
		\centering
		\subfloat[]{      
			\includegraphics[width=1\linewidth]{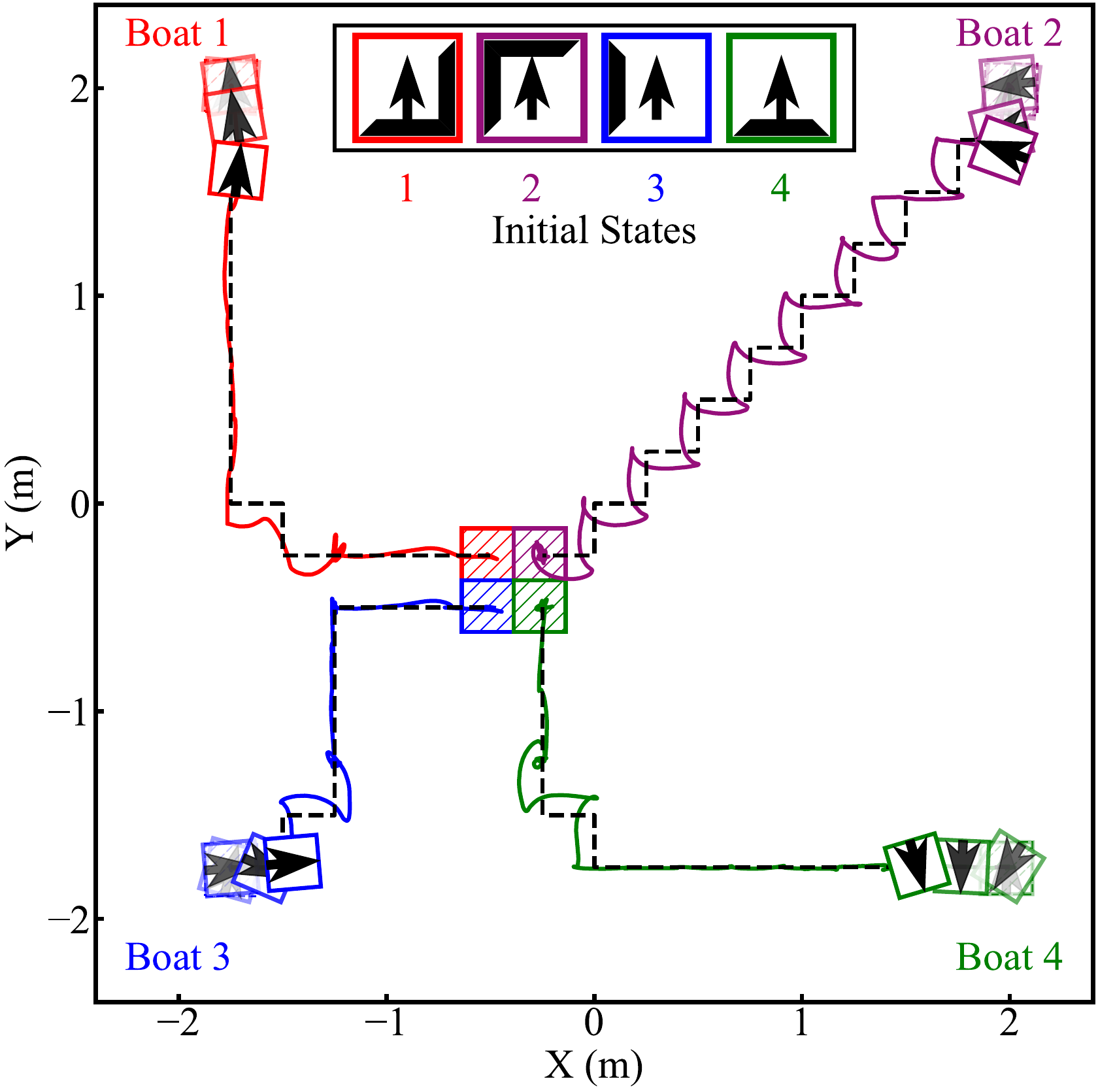}  
		}
	\end{minipage}
	\begin{minipage}{\pathfigwidth\linewidth}      
		\centering
		\subfloat[]{      
			\includegraphics[width=1\linewidth]{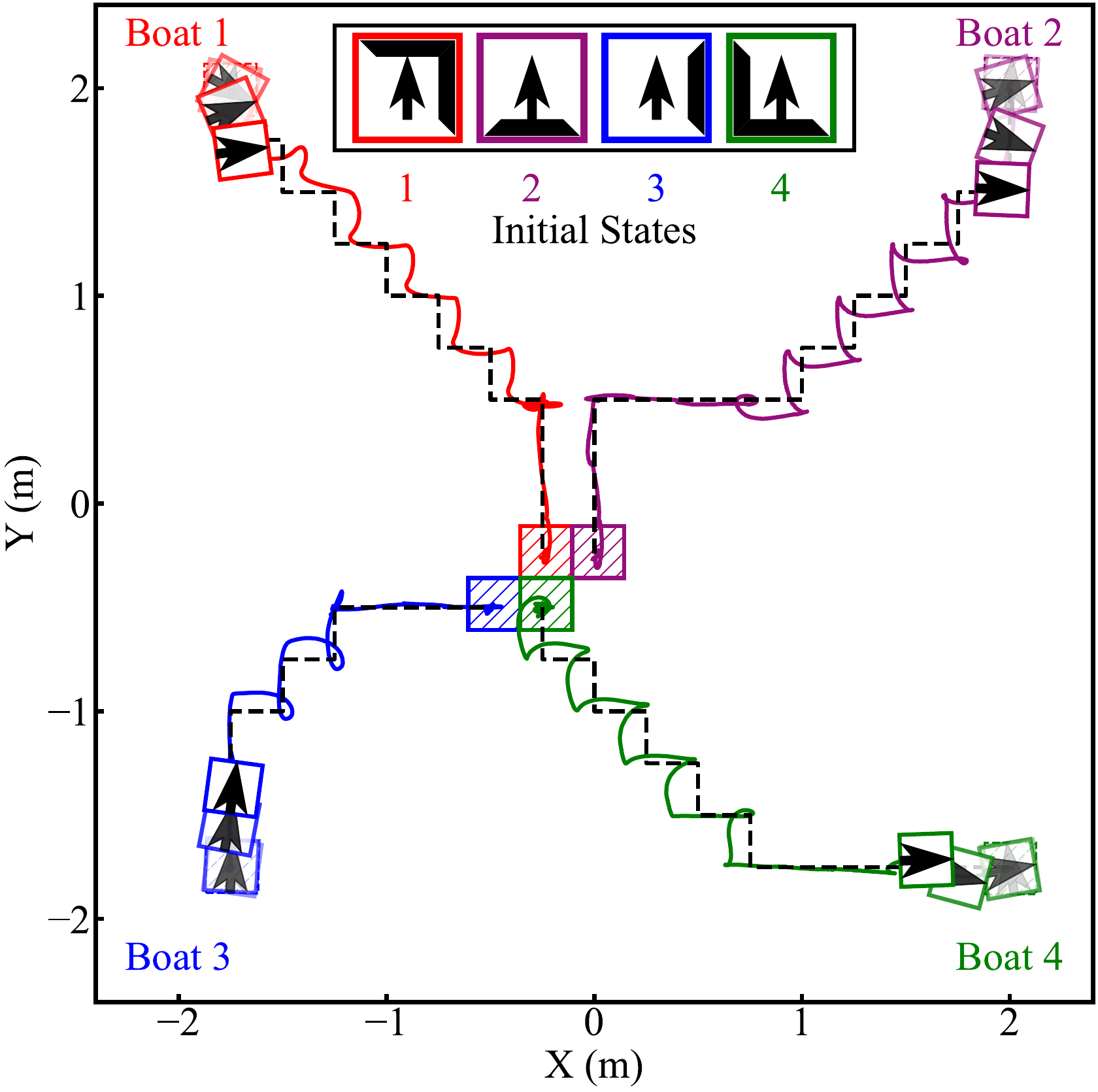}  
		}
	\end{minipage}
	\begin{minipage}{\pathfigwidth\linewidth}      
		\centering
		\subfloat[]{      
			\includegraphics[width=1\linewidth]{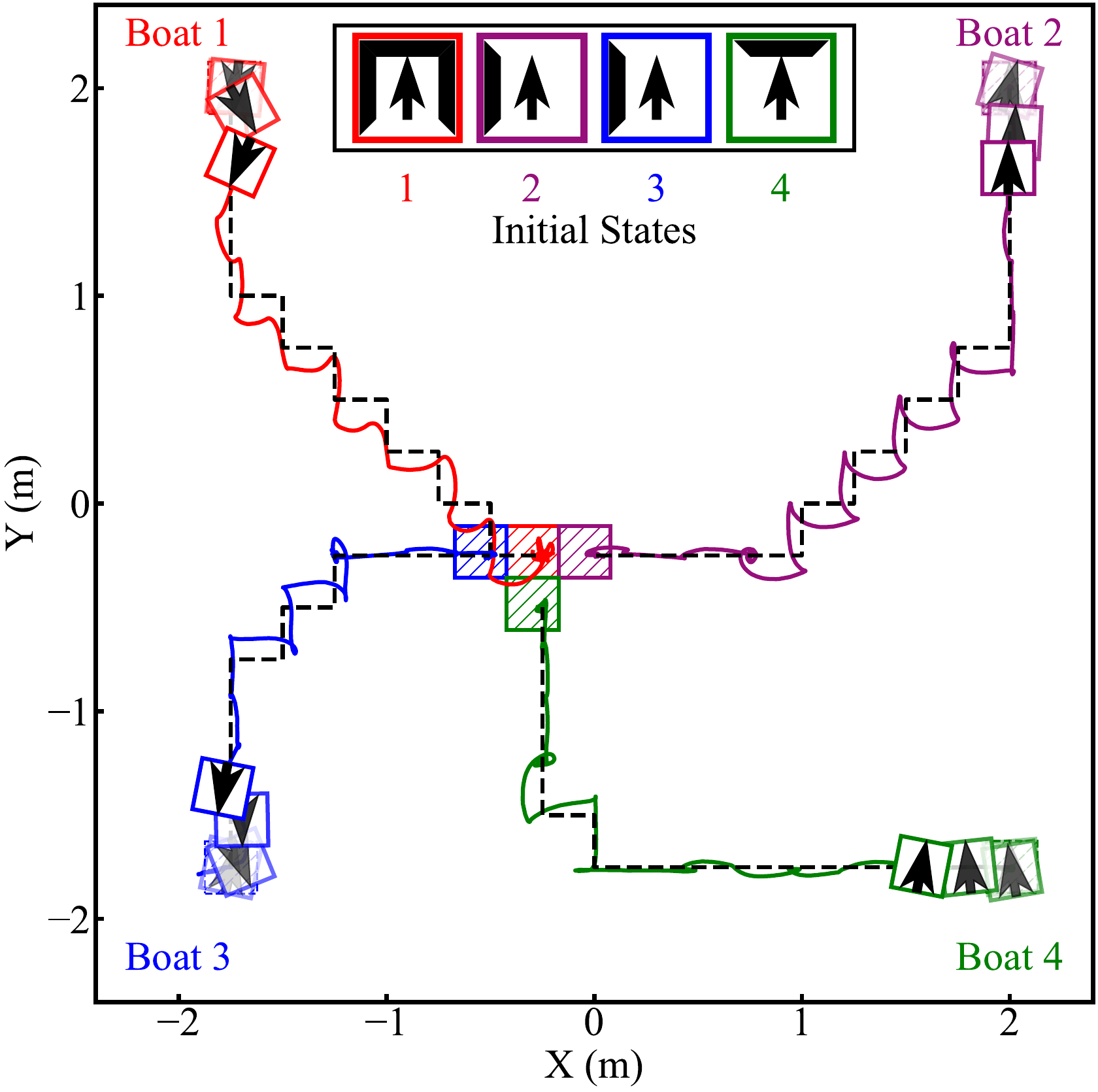}  
		}
	\end{minipage}
	
	\begin{minipage}{\pathfigwidth\linewidth}      
		\centering
		\subfloat[]{      
			\includegraphics[width=1\linewidth]{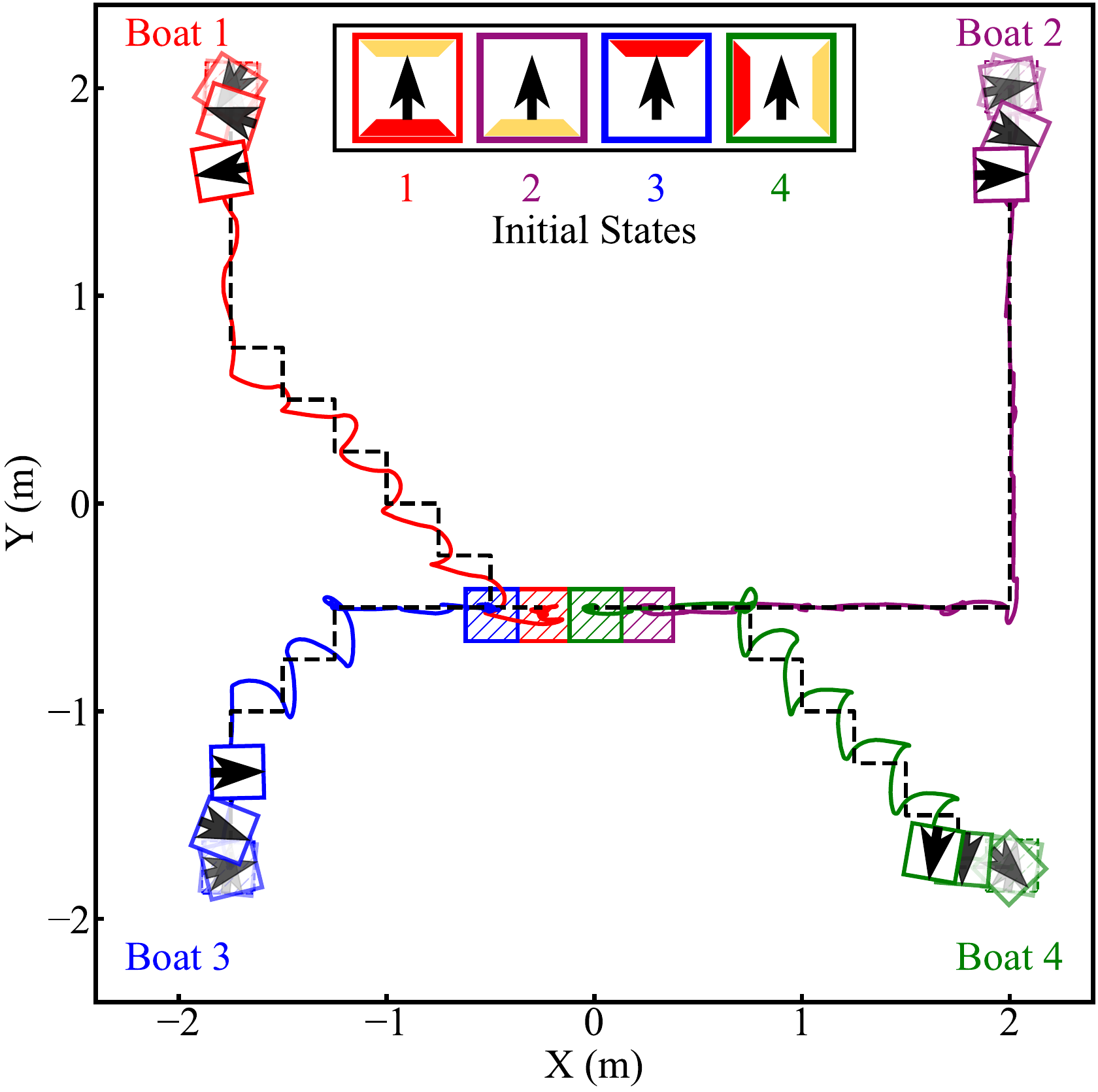}  
		}
	\end{minipage}
	\begin{minipage}{\pathfigwidth\linewidth}      
		\centering
		\subfloat[]{      
			\includegraphics[width=1\linewidth]{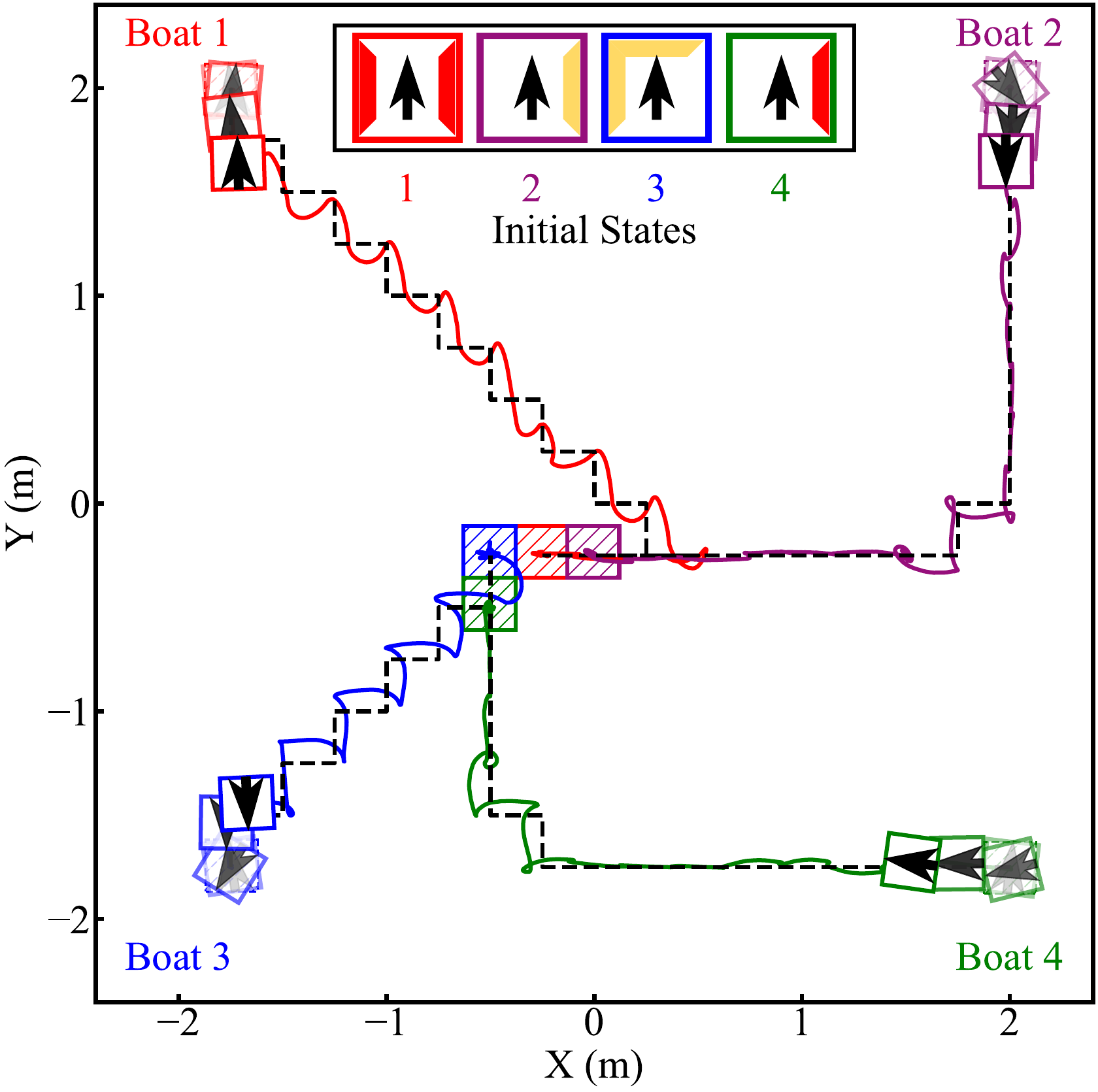}  
		}
	\end{minipage}
	\begin{minipage}{\pathfigwidth\linewidth}      
		\centering
		\subfloat[]{      
			\includegraphics[width=1\linewidth]{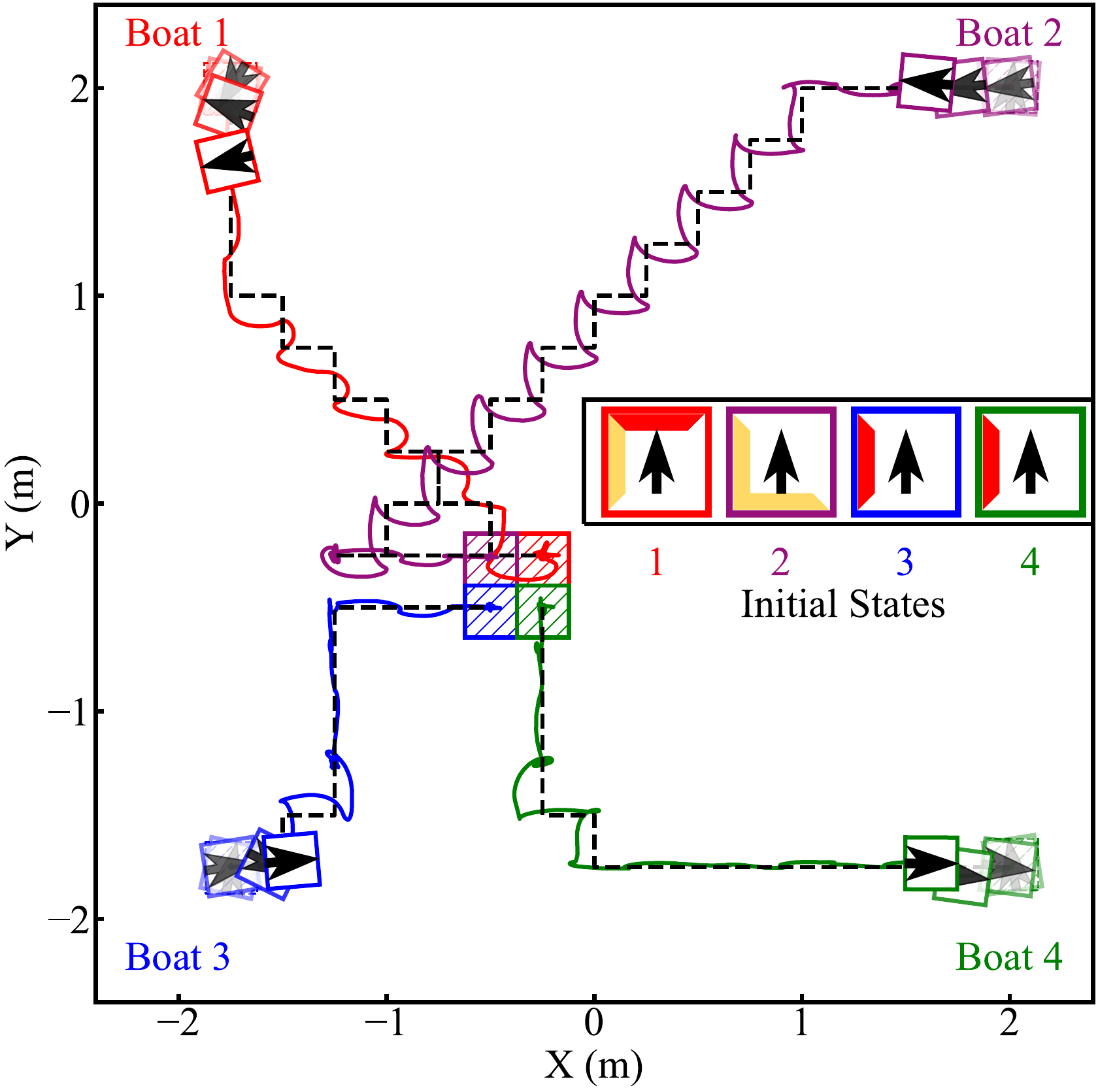}  
		}
	\end{minipage}
	\begin{minipage}{\pathfigwidth\linewidth}      
		\centering
		\subfloat[]{      
			\includegraphics[width=1\linewidth]{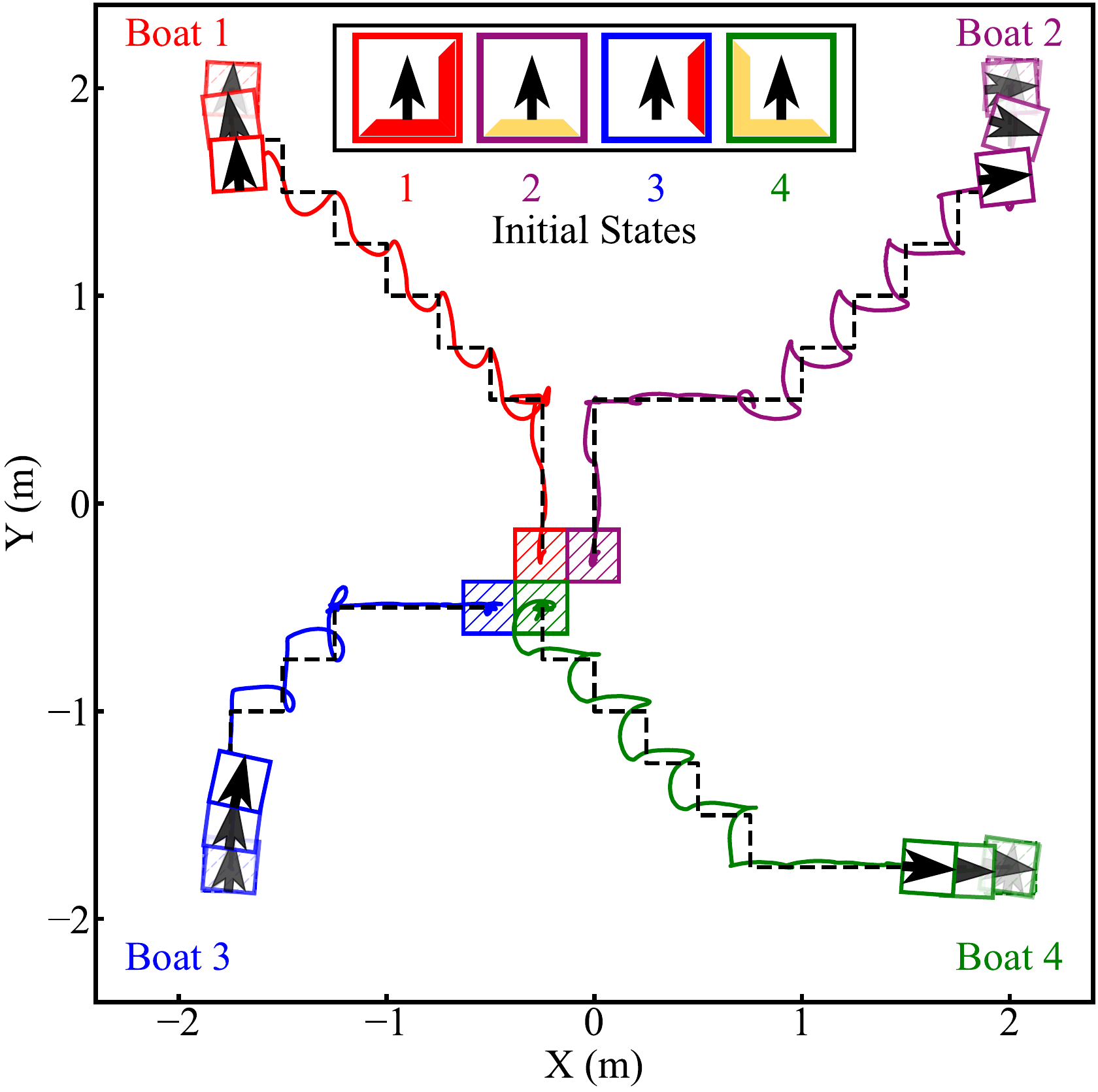}  
		}
	\end{minipage}
	\begin{minipage}{\pathfigwidth\linewidth}      
		\centering
		\subfloat[]{      
			\includegraphics[width=1\linewidth]{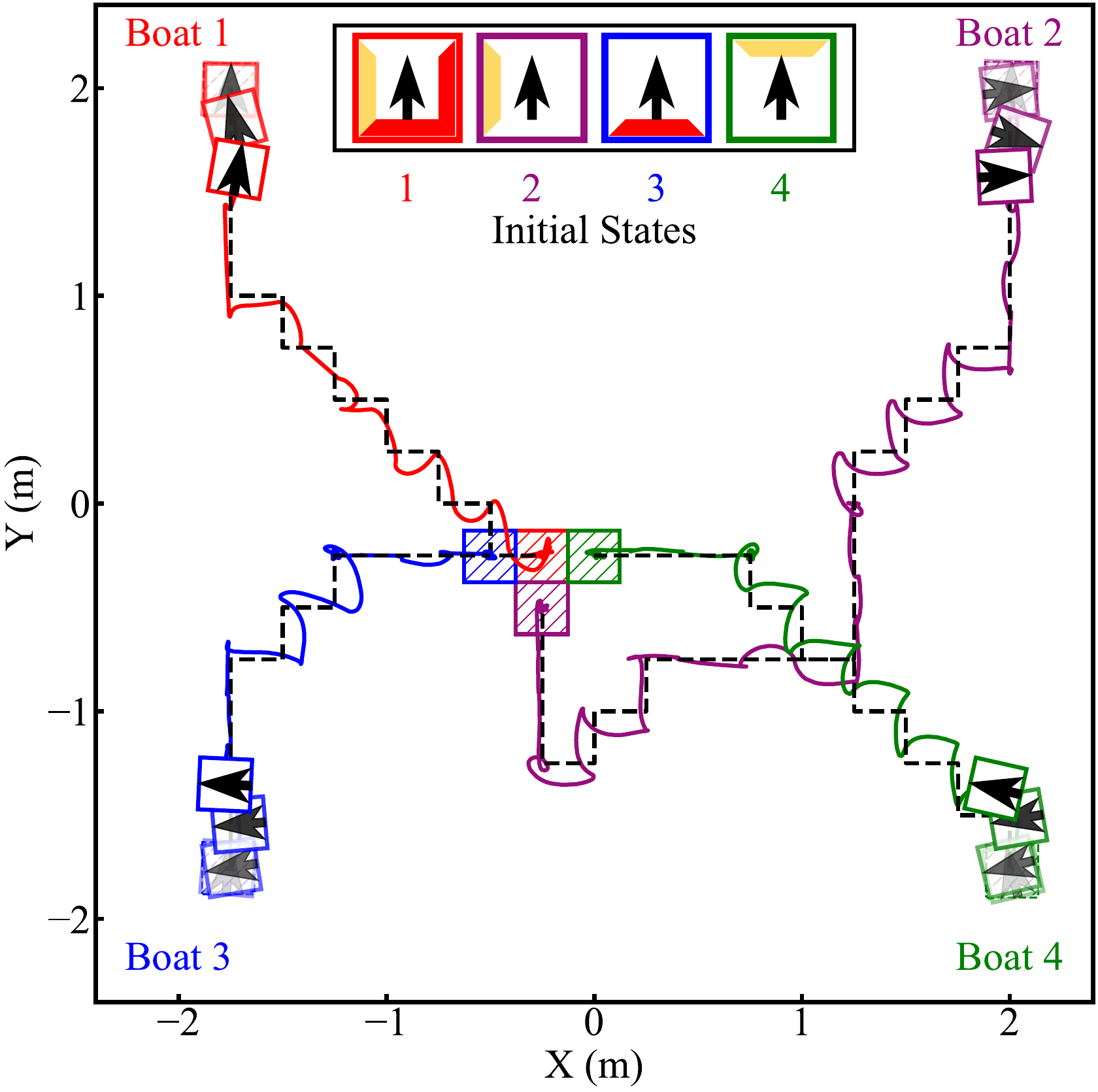}  
		}
	\end{minipage}

	\begin{minipage}{\pathfigwidth\linewidth}      
		\centering
		\subfloat[]{      
			\includegraphics[width=1\linewidth]{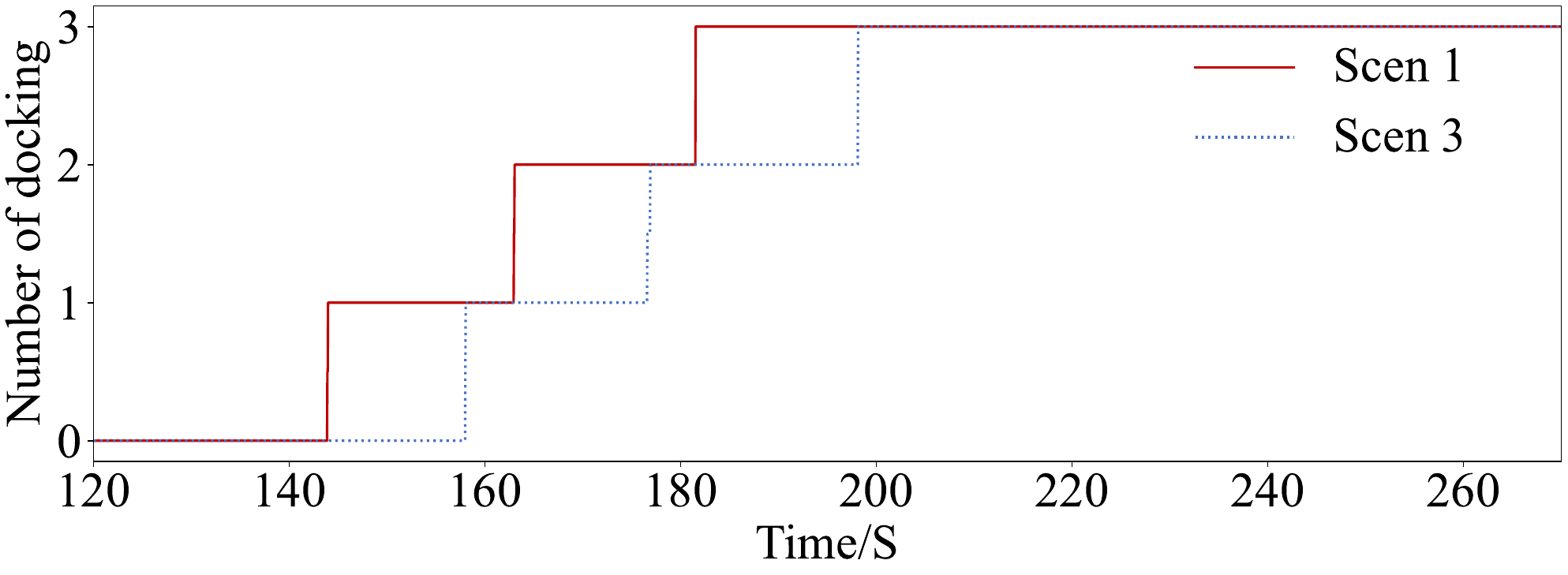}  
		}
	\end{minipage}
	\begin{minipage}{\pathfigwidth\linewidth}      
		\centering
		\subfloat[]{      
			\includegraphics[width=1\linewidth]{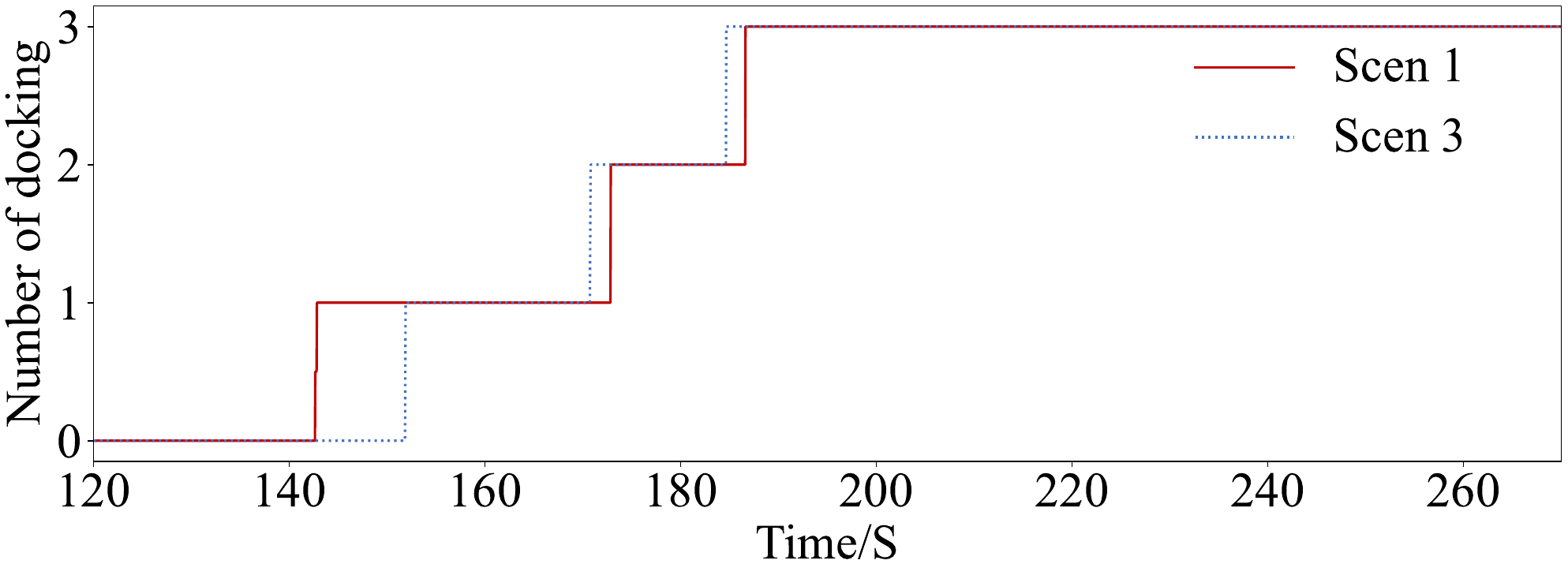}  
		}
	\end{minipage}
	\begin{minipage}{\pathfigwidth\linewidth}      
		\centering
		\subfloat[]{      
			\includegraphics[width=1\linewidth]{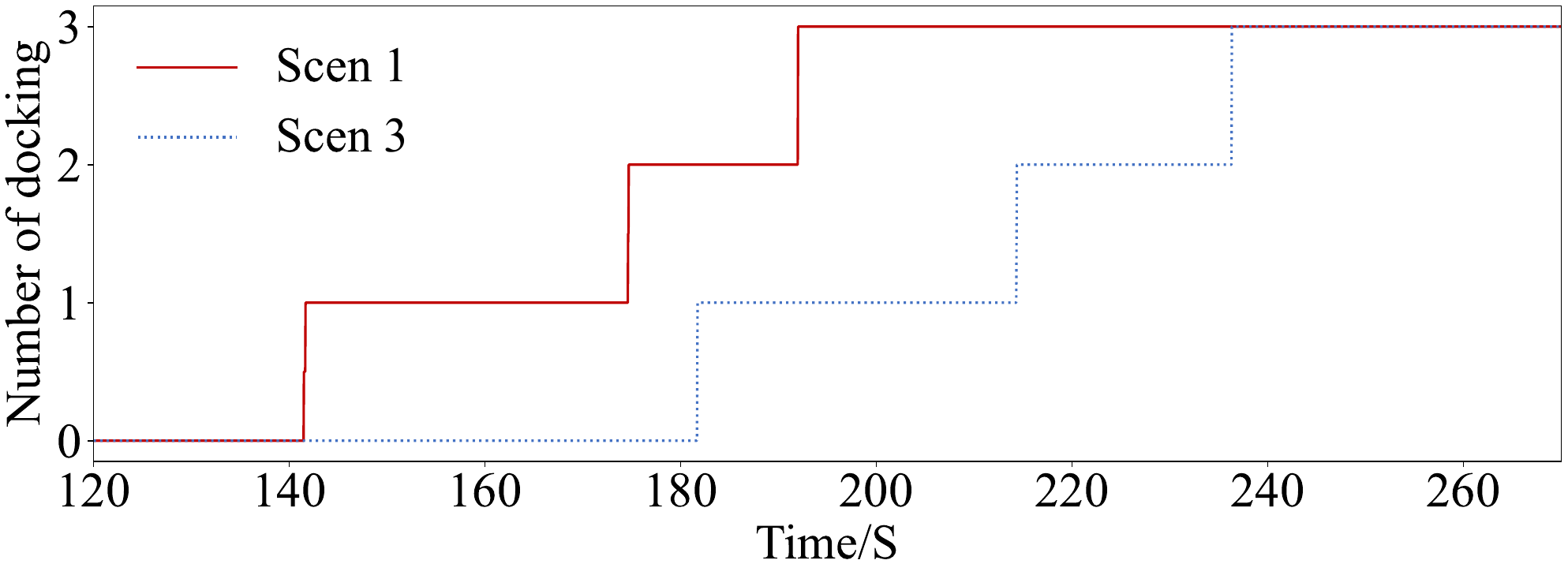}  
		}
	\end{minipage}
	\begin{minipage}{\pathfigwidth\linewidth}      
		\centering
		\subfloat[]{      
			\includegraphics[width=1\linewidth]{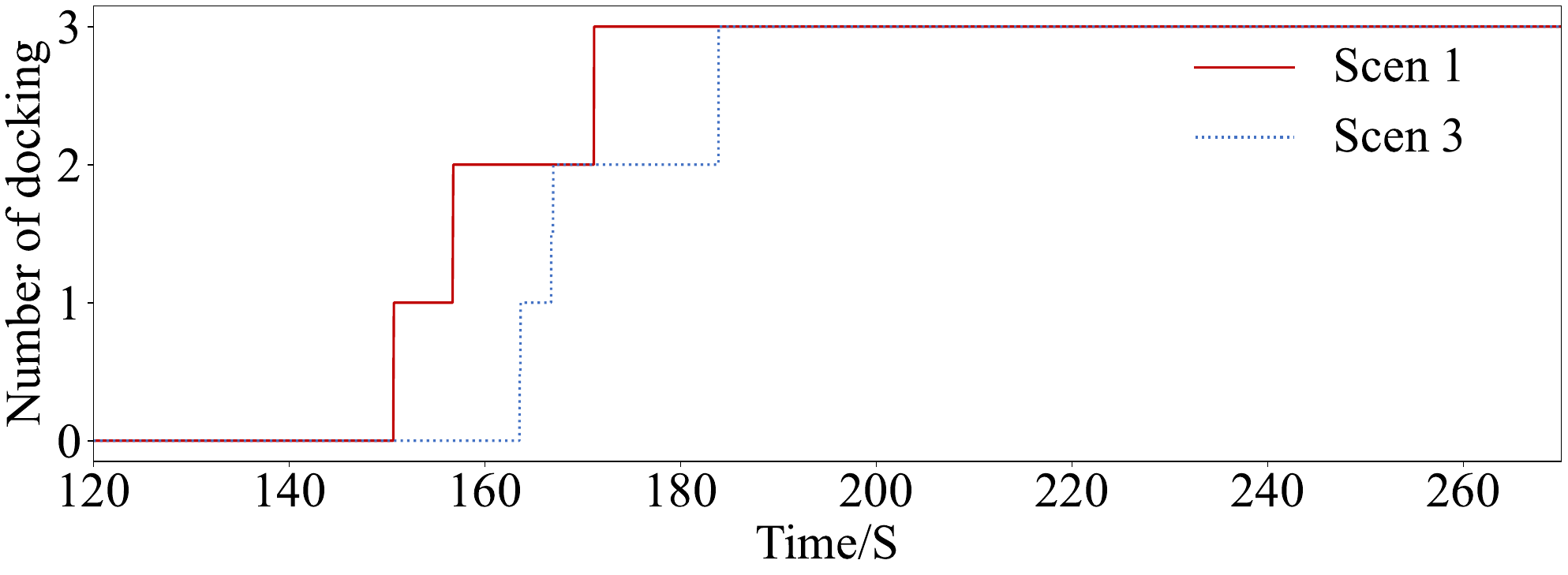}  
		}
	\end{minipage}
	\begin{minipage}{\pathfigwidth\linewidth}      
		\centering
		\subfloat[]{      
			\includegraphics[width=1\linewidth]{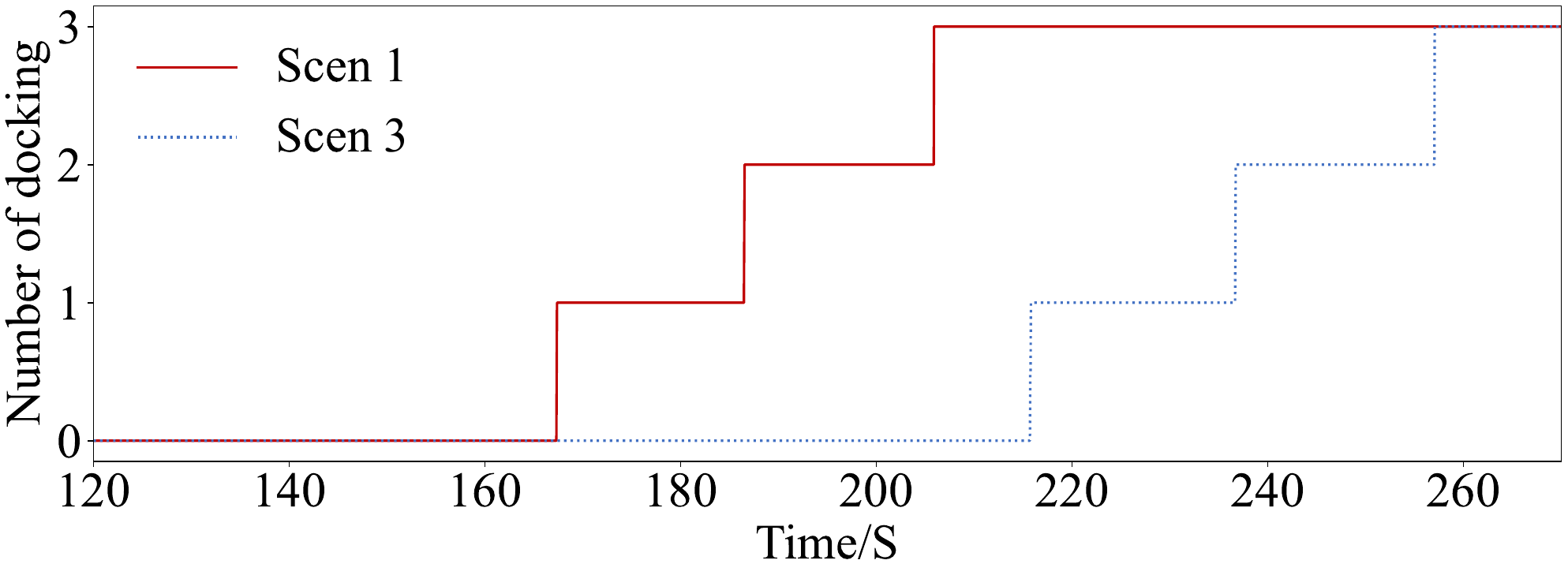}  
		}
	\end{minipage}
	
	\caption{ 
	Self-assembly experiments of four CuBoats with various docking system layouts: Genderless and Gender-opposite trials. Panels (a)-(e) and (f)-(j) illustrate the initial states and trajectories of the CuBoats in Genderless and Gender-opposite trials, respectively. 
	The color coding of robot edges in the Initial States boxes represents the genderless docking systems (black) and gendered docking systems (red-yellow).
	The boxes with arrows depict the CuBoats' orientation adjustments in the initial phase, while the boxes with shadows indicate their final destinations. The dashed lines represent the reference trajectories generated by the A* algorithm, and the continuous lines depict the experimental trajectories. Panels (k)-(o) display the evolution of completed connections over time.}
	\vspace{-10pt}
	\label{fig:exp_path}
\end{figure*}

We conducted five self-assembly experiments with different target structures for both the Genderless and Gender-opposite scenarios.
Figures \ref{fig:exp_path} (a)-(j) illustrate the trajectories of all CuBoats during these experiments. 
In both scenarios, the CuBoats initially rotate to their target directions in the first moves, ensuring the connectivity of the final desired structures. 
Subsequently, they simultaneously move towards their respective target regions with constant orientations, following the generated paths until two CuBoats are in close proximity and begin to assemble.
Throughout the assembly process, the groups of CuBoats always dock in pairs, adhering to the 1-to-1 docking mechanism. As a result, all desired structures are successfully assembled as expected.
Notably, all robots need only 6 docking mechanisms, much less than the 16 in systems with identical DMLs.

Figures \ref{fig:exp_path} (k)-(o) plot the evolution of the completed connections in both Genderless and Gender-opposite scenarios over time.
We can see that for the five maps, the completion time in the Gender-opposite scenario is generally longer than that in the Genderless scenario due to the limitation of gendered docking systems.
Upon observing the five maps, it becomes evident that the completion time in the Gender-opposite scenario is generally longer than that in the Genderless scenario due to the limitation imposed by the heterogeneous docking systems.
In certain instances, the CuBoats in the Gender-opposite scenario must take detours to execute the docking actions, leading to delayed completion. For example, Boat 2 in panels (h) and (j) demonstrates such a detour maneuver as it navigates towards its target position. Nevertheless, despite the occasional detours, the CuBoats eventually manage to establish connections.

\section{Conclusion}
\label{sect:conclusion}
In this paper, we introduce a multi-USV testbed system, featuring four omnidirectional USVs named CuBoats, each equipped with a heterogeneous docking system layout.
These CuBoats serve as dynamic platforms in inland rivers or harbors, facilitating various aquatic tasks.
To efficiently assemble CuBoats in a coordinated manner, we propose a generalized parallel SAP algorithm specifically tailored for this multi-USV system.
The SAP algorithm ensures the connectivity of the final constructed structure by proposing a dispatching approach that maps robots to target locations and generates an interconnected formation using the tabu search algorithm.
The algorithm provides target locations and orientations for dispatched robots in the first stage and generates an assembly tree in the second stage to determine the (dis)assembly sequence, ensuring connectivity within each node. This approach enables parallel movement and assembly of all dispatched modular CuBoats.
To demonstrate the practicality and efficacy of the proposed SAP algorithm, we conducted simulations on five maps with distinct target structures, testing the algorithm under two scenarios. 
Additionally, we performed real-world experiments involving four CuBoats on five maps, including both scenarios. The results from both simulations and experiments validate the reliability and efficiency of our algorithm, making it a valuable tool for autonomous robotic assembly in swarm robotics applications.






\bibliographystyle{IEEEtran}
\bibliography{selfassembly.bib}

\end{document}